\documentclass[manuscript]{acmart}

\usepackage{amsmath}
\usepackage{times}
\usepackage{graphicx}

\usepackage[outdir=./figures/]{epstopdf}
\usepackage{color}
\usepackage{multirow}
\usepackage{wasysym}

\usepackage{rotating}
\usepackage{bbm}
\usepackage{latexsym}

\usepackage{wrapfig}

\usepackage{subfigure}
\usepackage{multirow}
\usepackage{comment}

\usepackage{lipsum}
\usepackage{pifont}

\usepackage{color}
\usepackage{tikz}
\usepackage{bm}
\usepackage{times}
\usepackage[ruled,linesnumbered]{algorithm2e}
\usepackage[utf8]{inputenc}
\usepackage[english]{babel}
\usepackage{lineno}

\usepackage{amsthm}
\usepackage{url}
\usepackage{marvosym}

\usepackage{enumitem}

\usepackage{lipsum}
\usepackage[bottom]{footmisc}
\usepackage[utf8]{inputenc}
\usepackage[english]{babel}
\usepackage{booktabs}
\usepackage{makecell}
\newtheorem{theorem}{Theorem}

\newtheorem{lemma}{Lemma}
\usepackage{amsthm}
\usepackage{braket}

\usepackage{algorithm2e,setspace}
\usepackage{caption}
\usepackage{marvosym}
\newcommand{\ie}{\emph{i.e.}}
\newcommand{\eg}{\emph{e.g.}}
\newcommand*{\dif}{\mathop{}\!\mathrm{d}}

\def\cao{\textcolor{black}}

\usepackage{thmtools}
\declaretheoremstyle[headfont=\normalfont]{normalhead}

\newtheorem{proposition}{Proposition}

\newtheorem{insight}{Insight}

\DeclareMathOperator*{\argmin}{arg\,min}

\AtBeginDocument{%
  }

\setcopyright{acmlicensed}
\copyrightyear{2018}
\acmYear{2018}
\acmDOI{XXXXXXX.XXXXXXX}
\acmConference[Conference acronym 'XX]{Make sure to enter the correct
  conference title from your rights confirmation email}{June 03--05,
  2018}{Woodstock, NY}
\acmISBN{978-1-4503-XXXX-X/2018/06}

\begin{document}

\title{Analytical Survey of Learning  with Low-Resource Data: From Analysis to Investigation}

\author{Xiaofeng~Cao}
\email{xiaofeng.cao.uts@gmail.com}
\affiliation{%
  \institution{School of Computer Science and Technology, Tongji University}
  \city{Shanghai}
  \country{China}
  \postcode{201804}
}

\author{Mingwei~Xu}
\email{mingweixu20@outlook.com}
\affiliation{%
  \institution{School of Artificial Intelligence, Jilin University}
  \city{Changchun}
  \country{China}
  \postcode{130012}
}

\author{Xin~Yu}
\email{xin.yu@uq.edu.au}
\affiliation{%
  \institution{School of Electrical Engineering and Computer Science, The University of Queensland}
  \city{Brisbane}
  \country{Australia}
}

\author{Jiangchao~Yao}
\email{Sunarker@sjtu.edu.cn}
\affiliation{%
  \institution{Cooperative Medianet Innovation Center, Shanghai Jiao Tong University}
  \city{Shanghai}
  \country{China}
}

\author{Wei~Ye}
\email{yew@tongji.edu.cn}
\affiliation{%
  \institution{College of Electronic and Information Engineering, Tongji University}
  \city{Shanghai}
  \country{China}
}

\author{Shengjun~Huang}
\email{huangsj@nuaa.edu.cn}
\affiliation{%
  \institution{ College of Computer Science and Technology, Nanjing University of Aeronautics and Astronautics}
  \city{Nanjing}
  \country{China}
}

\author{Minling~Zhang}
\email{zhangml@seu.edu.cn}
\affiliation{%
  \institution{School of Computer Science and Engineering, Southeast University}
  \city{Nanjing}
  \country{China}
}

\author{Ivor W. Tsang}
\email{ivor_tsang@cfar.a-star.edu.sg}
\affiliation{%
  \institution{ Centre for Frontier AI Research and Institute of High Performance Computing, Agency for Science, Technology and Research (A$^*$STAR)}
  \country{Singapore}
}

\author{Yew Soon Ong}
\email{asysong@ntu.edu.sg}
\affiliation{%
  \institution{School of Computer Science and Engineering, Nanyang Technological University, and Agency for Science, Technology and Research (A$^*$STAR)}
  \country{Singapore}
}

\author{James T. Kwok}
\email{jamesk@cse.ust.hk}
\affiliation{%
  \institution{Department of Computer Science and Engineering, The Hong Kong University of Science and Technology}
  \city{Hong Kong}
  \country{Hong Kong SAR}
}
  \author{Heng~Tao Shen\textrm{\textsuperscript{\Letter}}}
\email{shenhengtao@hotmail.com}
\affiliation{%
  \institution{School of Computer Science and Technology, Tongji University}
  \city{Shanghai}
  \country{China}
  \postcode{201804}
}
\renewcommand{\shortauthors}{Xiaofeng Cao,  et al.}

\begin{abstract}
Learning with high-resource data has demonstrated substantial success in artificial intelligence (AI); however, the costs associated with data annotation and model training remain significant. A fundamental objective of AI research is to achieve robust generalization with limited-resource data. This survey employs agnostic active sampling theory within the Probably Approximately Correct (PAC) framework to analyze the generalization error and label complexity associated with learning from low-resource data in both model-agnostic supervised and unsupervised settings. Based on this analysis, we investigate a suite of optimization strategies tailored for low-resource data learning, including gradient-informed optimization, meta-iteration optimization, geometry-aware optimization, and LLMs-powered optimization. Furthermore, we provide a comprehensive overview of multiple learning paradigms that can benefit from low-resource data, including domain transfer, reinforcement feedback,  and hierarchical structure modeling.  Finally, we conclude our analysis and investigation by summarizing the key findings and highlighting their implications for learning with low-resource data.
\end{abstract}



\keywords{Low-resource data, active sampling, PAC framework, theoretical guarantee, model-agnostic, hyperbolic, domain transfer, reinforcement feedback,  and hierarchical structure modeling.}
\maketitle

\section{Introduction}
That's a cat sleeping in the bed, the boy is patting the elephant, and those are people going on an airplane. That's a big airplane..." "This is a three-year-old child describing the pictures she sees," said Fei-Fei Li. She presented a famous lecture on "How We Are Teaching Computers to Understand Pictures" at the Technology Entertainment Design (TED) 2015 event \footnote{\url{https://www.ted.com/talks/fei_fei_li_how_we_re_teaching_computers_to_understand_pictures?language=en}}.
In the real world, humans can recognize objects and scenarios simply by relying on a single picture based on their prior knowledge \cite{dicarlo2012does}. Machines, on the other hand, may require more. Over the past few decades, artificial intelligence (AI) \cite{russell2002artificial,nilsson2014principles}, as a branch of computer science, mathematics, and physics, has aimed to develop automated processes that can partially or fully mimic human intelligence. Through machine learning techniques, particularly its subset deep learning, AI systems have significantly enhanced their intelligence by learning from high-resource data \cite{o2013artificial,labrinidis2012challenges}, approaching human-like intelligence in certain aspects. For instance, by modeling the neuron propagation mechanisms of the human brain, a series of highly effective AI systems have been developed, such as Deep Blue \cite{campbell2002deep} and AlphaGo \cite{wang2016does}.

Of course, the talent of AI is not innate. Training from high-resource data helps AI recognize different objects and scenarios\cite{lecun2015deep}. Recent advancements in large language models (LLMs) have significantly optimized the utilization of high-resource data, leading to substantial improvements in the efficacy and performance of model training processes. For instance, models like ChatGPT\cite{brown2020language}, LLaMA\cite{touvron2023llama}, Google's PaLM\cite{anil2023palm}, and DeepSeek\cite{bi2024deepseek} have demonstrated remarkable capabilities in natural language understanding and generation. However, training and annotating large-scale data can be quite expensive, even when we adopt machine annotation techniques.

“AI is not just for the big guys anymore\footnote{\url{https://www.forbes.com/sites/forbestechcouncil/2020/05/19/the-small-data-revolution-ai-isnt-just-for-the-big-guys-anymore/?sh=4b8d35c82bbb}}.” A novel perspective suggests that the low-resource data revolution is ongoing, and training with low-resource data to achieve desired performance is one of the ultimate purposes of AI \cite{longpre2024consent}. For example, we can fine-tune LLMs with a limited set of examples \cite{radford2018improving,MoA-VR,VARFVV}, thereby rendering the utilization of low-resource data both practical and effective\footnote{\url{https://www.superannotate.com/blog/llm-fine-tuning}}.
Technically, human experts expect to reduce reliance on high-resource data and discover new breakthroughs for AI systems, especially in the deployment and configuration of deep neural networks \cite{segler2018planning}. Researchers in low-resource deep learning have addressed data scarcity through: (1) training with limited labeled data using semi-supervised or weakly supervised learning methods \cite{chen2019scene,iosifidis2017large,hong2025dual}; (2) developing models that perform effectively with fewer labeled examples, employing techniques such as few-shot, transfer, or active learning \cite{luvcic2019high,ji2019learning,xu2020weakly}; and (3) training on smaller datasets by incorporating data augmentation or leveraging prior knowledge \cite{baraniuk2011more,moore1993prioritized}. Formally, few-shot learning \cite{sung2018learning,hedderich2020survey,song2023comprehensive}, often referred to as low-resource data learning, is a unified topic that studies low-resource data with limited information. Based on the work of \cite{wang2020generalizing}, an explicit scenario of few-shot learning is feature generation \cite{xian2019f}, which involves generating artificial features from the given limited or insufficient information.
Another scenario, which involves implicit supervision information, is more challenging and relies on retraining the learning model \cite{xian2019f,feng2019partial} with highly informative examples, such as private data.

\textbf{Questions Behind the Investigation.} \emph{Theoretically, in most few-shot learning scenarios, the label distribution is passive, meaning it is explicitly controlled by finite training resources based on known distributions. As an ambitious and aggressive learning paradigm, it aims to achieve consistent generalizations using limited training samples from predefined distributions. However, the feasibility of achieving such generalizations remains challenging. Therefore, active learning, as defined by \cite{settles2009active}, draws our attention because it involves label acquisitions controlled by a learning algorithm or human input.}

\textbf{Foundation of  the Analytical Survey.}  Different from few-shot learning, the annotation scenario of active learning is not as restricted; it offers more flexibility in terms of training resource configurations. Typically, an active learning algorithm can halt its iterative sampling at any time based on desired algorithm performance or an exhausted annotation budget. There are two categories for active learning: active sampling theory over hypothesis classes \cite{hanneke2014theory} and active sampling algorithms over realized scenarios \cite{settles2012active}. The theory studies present the label complexity and convergence guarantees for these algorithmic paradigms.  Typically, their   analyses are derived from a PAC (Probably Approximately Correct) \cite{haussler1990probably} style, aiming for an agnostic setting such as \cite{dasgupta2008general}.
To control active sampling, there is a type of error disagreement coefficient that identifies target data, which can maximize the hypothesis updates most effectively; these updates are required to be positive and helpful. Therefore, active sampling is also a hypothesis-pruning \cite{cao2020shattering} process that seeks to find the optimal hypothesis from a given hypothesis class, maintaining hypotheses from a version space \cite{hirsh1994generalizing, mitchell1977version} over the decision boundaries \cite{lee1997decision} of classes. 

\subsection{Motivations and Contributions}
In the future, leveraging low-resource data will be essential for advancing artificial intelligence. As a proactive research area, few-shot learning has emerged as a significant approach for training with limited data, particularly in fine-tuning large language models (LLMs). While few-shot learning operates within a passive framework that often provides insufficient label information for the task at hand. Additionally, there are few theoretical guarantees regarding its generalization performance in task-independent settings. This gap motivates us to offer a theoretical analysis for learning with low-resource data, focusing on model-agnostic generalization.

\begin{figure} 
  \centering  
  \includegraphics[width=0.95\textwidth]{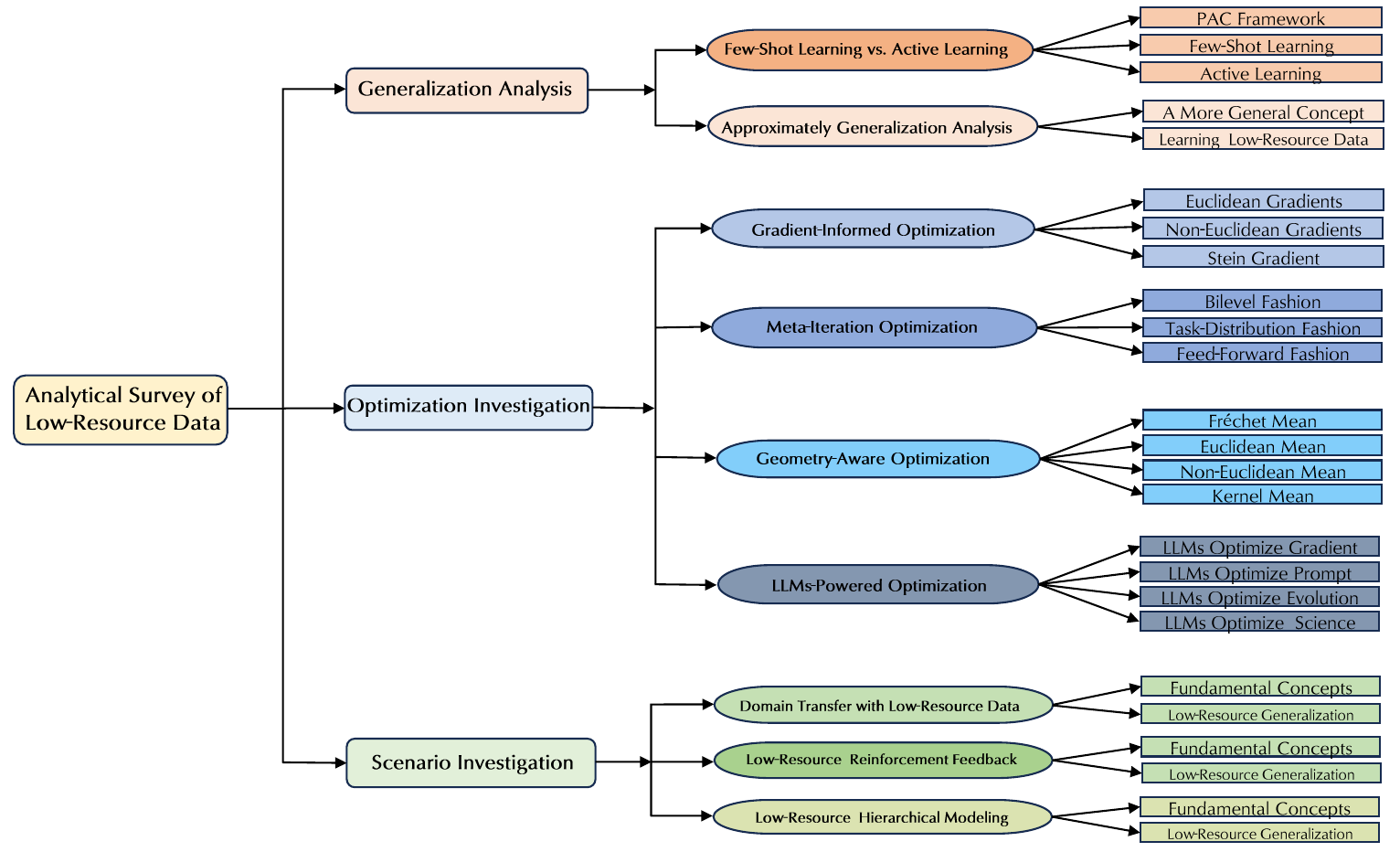}  
  \caption{The key contents of \emph{Analytical Survey of Learning with Low-Resource Data: From Analysis to Investigation}.}  
  \Description{The key contents}
  \label{fig:Low-Resource Learning} 
\end{figure}

Recognizing that active learning theory offers an effective i.i.d. sampling protocol, we adopt its PAC framework to establish error and label complexity bounds for learning in low-resource data settings. This analysis initiates our investigation into optimization strategies and scenarios pertinent to low-resource data. In Figure \ref{fig:Low-Resource Learning}, we present the key content of this paper. The contributions of this survey are summarized as follows:

\begin{itemize}
\item We present a formal analysis for learning with low-resource data, emphasizing efficient generalization approximations to high-resource data representations. This analysis operates in a model-agnostic framework, deriving a more shallow concept from a generalization protocol.

\item From a PAC learning perspective, we present, to the best of our knowledge, the first theoretical guarantees for learning with low-resource data through the framework of active sampling theory. Specifically, we derive bounds on generalization error and label complexity within a model-agnostic context. Our findings indicate that learning from low-resource data can yield effective performance as high-resource data while maintaining an acceptable level of loss.

\item From a structural standpoint, we present several approaches such as gradient-informed optimization, meta-iteration optimization, geometry-aware optimization, and LLMs-powered optimization, that are crucial for effectively utilizing low-resource data. We specifically emphasize the importance of representing Euclidean and non-Euclidean geometries and the associated optimization solutions.

\item We investigate several learning paradigms, including domain transfer, reinforcement learning, and hierarchical structure modeling, to enhance data efficiency through low-resource data representation. This exploration seeks to improve the effectiveness of these methods in scenarios where data availability is limited or training costs are high, ultimately enabling more robust performance in resource-constrained scenarios.
\end{itemize}

\section{Analysis from Generalization}
\subsection{Few-Shot Learning vs. Active Learning}

Few-shot learning can be regarded as a proactive approach to learning with low-resource data with limited resources. In contrast, active learning also offers solutions for low-resource data, but with greater flexibility in sample selection. This distinction lies in the approach to resource collection, particularly concerning labels. Below, we compare these two strategies within the framework of  PAC  hypothesis theory.

\subsubsection{PAC Framework}
In 1984,  Leslie Valiant proposed a computational learning concept-Probably Approximately Correct (PAC) \cite{valiant1984theory}. PAC   presents  mathematical analysis for machine learning under fixed  distribution and parameter assumptions.   The learner is theoretically  
 required to select a generalization hypothesis (also called conceptual function) from a candidate hypothesis class by observing the received data and labels. The goal is to converge the hypothesis into approximately correct generalization that properly describes the probability distribution of the unseen samples. One key content of PAC leaning is to derive the computational complexity  bounds such as sample complexity \cite{hanneke2016optimal}, 
generalization errors, and Vapnik–Chervonenki (VC) dimension (the capacity of a classification model) \cite{blumer1989learnability}.

In computational learning theory, active learning attempts to prune the candidate infinite concept class into the optimal hypothesis, which retains consistent properties with its labeled examples \cite{hanneke2014theory}. The main difference from typical PAC learning is that active learning controls the hypothesis pruning by using fewer training data. Therefore, active learning, which finds hypotheses consistent with a small set of labeled examples, can be regarded as a standard hypothesis pruning approach in PAC learning. Within this framework, we present this survey.

\subsubsection{Few-Shot Learning}
Finding the optimal hypothesis consistent with the full training set is a standard theoretical description of machine learning in PAC. The convergence process involves performing hypothesis pruning within the candidate hypothesis class/space. Therefore, the number of hypotheses within a fixed geometric region determines the volume of the hypothesis space, which in turn affects the speed and cost of hypothesis pruning.

Given a full training set $\mathcal{X}$ with $n$ samples, let $\mathcal{H}$  be the hypothesis class, the VC dimension bound of $\mathcal{H}$ can be used to describe the complexity of the  convergence difficulty of the hypothesis-pruning 
for  a given   learning algorithm.
We thus follow the agnostic active learning \cite{dasgupta2008general} to define $N(\mathcal{H}, n, k, \mathcal{A})$ as a class of function which controls the convergence of  the learning algorithm  $\mathcal{A}$  in hypothesis class $\mathcal{H}$,  associating from  $n$ training samples with $k$  classes.  The following insight statements analyze each learning process through the lens of pruning within a model-agnostic hypothesis space. Comprehensive surveys on few-shot learning and active learning from a general data perspective have been conducted by Wang et al. \cite{wang2020generalizing} and Settles \cite{settles2009active}, respectively. Therefore, we do not elaborate further on these topics here.

\begin{insight}
Properties of Hypotheses in Machine Learning. From a hypothesis-pruning perspective, given any machine learning algorithm $\mathcal{A}$, its candidate hypothesis class  is characterized  by $\mathcal{H}$, which satisfies: 1) a VC dimension bound of  $\mathcal{O}(2^n)$, and 2) a safety uniform bound  of pruning into a non-null hypothesis is  $N(\mathcal{H}, n, k, \mathcal{A}) \geq \mathcal{O}\left(\frac{k-1}{k}n\right)$,  and 3)  the  VC dimension bound of a non-null hypothesis subspace is $\mathcal{O}\left(2^n - 2^{\frac{(k-1)n}{k}}\right)$.
\end{insight}

Note that the uniform bound aims at an expected complexity and a non-null hypothesis requires the training examples to cover all label categories. Given any class has at least $\eta$ data, a safety guarantee requires   $N(\mathcal{H}, n, k, \mathcal{A})\geq \mathcal{O}(n-\eta)$. For a uniform estimation on $\eta\approx \mathcal{O}(\frac{n}{k})$ over all possibilities of $\eta=1,2,3,...,\frac{n}{k}$, a safety uniform bound  of pruning into a non-null hypothesis is  $\mathcal{O}(\frac{k-1}{k}n)$.  

Assume that  $\eta \ll \frac{n}{k}$, the typical machine learning scenario becomes a few-shot learning process.
\begin{insight}\label{few-shot-learning}
Properties of Hypotheses in Few-shot Learning. From a hypothesis-pruning perspective, given any few-shot learning algorithm   
$\mathcal{A}$,   its candidate hypothesis class  is characterized  by $\mathcal{H}$, which satisfies   1) a VC dimension bound of  $\mathcal{O}(2^n)$,  2) a safety uniform bound  of pruning into a non-null hypothesis is  $N(\mathcal{H}, n, k, \mathcal{A})\geq \mathcal{O}(n-\eta)$, and 3) a  VC dimension bound of shrinking into a non-null hypothesis subspace is  $\mathcal{O}(2^n-2^{n-\eta})$, where $2^n-2^{n-\eta} \gg \frac{k-1}{k}n$.
\end{insight}

 Here, 'safety' means that the pruning process can properly converge.  From Insight~\ref{few-shot-learning}, few-shot learning can be deemed as a special case of typical machine learning with limited supervision information. One important characteristics of it is its tighter volume of the non-null hypothesis space since $\mathcal{O}(2^n-2^{n-\eta}) \gg \mathcal{O}(\frac{k-1}{k}n)$. Therefore, compared with typical machine learning, any few-shot learning algorithm  will result in looser safety bound to prune into a non-hypothesis. 

 In realizable settings,  one typical few-shot learning   scenario is    feature generation \cite{xian2019f} via model retraining. In this scenario, the learning algorithm generates  handwritten
features like humans by pre-training the model with prior knowledge, where the  retaining will not stop, until a desired performance is achieved.
However, retraining the learning model is adopted in rare cases, which cannot properly generate highly-trust features. For example,  one special case of few-shot learning is one-shot learning \cite{vinyals2016matching} which relies on only one data in some classes; the other  more extreme  case is zero-shot learning \cite{romera2015embarrassingly} where some classes do not include any data or labels.  Their detailed hypothesis properties are presented as follows.
\begin{insight}
  Properties of Hypotheses in One-shot Learning. 
  From a hypothesis-pruning perspective, given any one-shot learning algorithm $\mathcal{A}$, its candidate hypothesis class is characterized by $\mathcal{H}$,  which satisfies: 1) a VC dimension of $\mathcal{H}$ is bounded by $\mathcal{O}(2^n)$,  2) a safety uniform bound for pruning to a non-null hypothesis is $N(\mathcal{H}, n, k, \mathcal{A})\geq \mathcal{O}(n)$, where $n=k$, 3) a VC dimension of any non-null hypothesis subspace is bounded by $\mathcal{O}(2^k)$.  
\end{insight}
 
Different to the 'safety' bound, 'inappropriate' means that the pruning can not converge into the desired hypothesis.
 
\begin{insight}
Properties of Hypotheses in Zero-shot Learning. From a hypothesis-pruning perspective, given any zero-shot learning algorithm   
$\mathcal{A}$,   its candidate hypothesis class  is characterized  by $\mathcal{H}$, which satisfies   1) a VC dimension bound of  $\mathcal{O}(2^n)$,  2)  an inappropriate safety uniform bound  of pruning into a non-null hypothesis, and 3) an inappropriate  VC dimension bound of shrinking into a non-null hypothesis subspace.
\end{insight}
  
Usually, the few-shot learning is also related to weakly-supervised learning \cite{liu2019prototype,li2019towards}, which includes incomplete, inaccurate, noisy, and outlier information, \emph{etc.} From this perspective, few-shot learning can be considered as one special setting of weakly-supervised learning with incomplete label information.
Imbalanced learning \cite{he2009learning}, transfer learning \cite{yu2020transmatch}, meta-learning \cite{jamal2019task}, \emph{etc.}, also have inherent connections with few-shot learning. However, there are no theoretical analyses for the convergence of the optimal hypothesis.

\subsubsection{Active Learning} 
Active learning \cite{settles2009active,ren2021survey} prunes the candidate hypothesis class to achieve a desired convergence. The pruning process typically involves zooming out the hypothesis space by querying highly informative updates \cite{cortes2019active}. Therefore, the assumption in active learning is that any update resulting from hypothesis pruning should be non-null. Here, we describe the hypothesis properties of active learning.

\begin{insight}
Properties of Hypotheses in Active  Learning. From a hypothesis-pruning perspective, given any active  learning algorithm   
$\mathcal{A}$ with a querying budget $\mathcal{Q}$,   its candidate hypothesis class  is characterized  by $\mathcal{H}$, which satisfies   1) a VC dimension bound of  $\mathcal{O}(2^n)$,  2) a safety uniform bound  of pruning into a non-null hypothesis is   $\mathcal{O}(1)$, and 3) the VC dimension bound of a non-null hypothesis subspace is   $\mathcal{O}(2^\mathcal{Q})$.
 \end{insight}

Note that active learning requires the updates of hypotheses are positive and non-null. Any subsequent hypotheses can converge into a safety state, which derives a safety uniform bound of $\mathcal{O}(1)$. 
Different from  few-shot learning, the scenario  of active learning is controlled by humans, which always keep a non-null update on hypotheses. It is thus that its VC dimension bound is tighter than the typical machine learning and few-shot learning. To find feasible hypothesis updates, active learning always uses an error disagreement coefficient  \cite{dasgupta2011two} to control the hypothesis-pruning.

\textbf{Error disagreement.} Given a finite hypothesis class $\mathcal{H}$, active learning iteratively updates  the current hypothesis $h_\mathcal{Q}\in \mathcal{H}$  at $t$-time into the optimal hypothesis $h^*\in \mathcal{H}$. Let an active learning algorithm $\mathcal{A}$ perform $\mathcal{Q}$ rounds of querying from $\mathcal{X}$,  assume that $\ell(\cdot,\cdot)$ denotes the loss of mapping $\mathcal{X}$ into $\mathcal{Y}$ with multi-class setting, we define the total loss of the  $\mathcal{Q}$ rounds of querying as $R(h_\mathcal{Q})=\sum_{i=1}^{\mathcal{Q}}\frac{q_t}{p_i}\ell(h(x_t), y_t)$, where $y_t$ denotes the label of $x_t$, $q_t$ satisfies the Bernoulli distribution of $q_t\in \{0,1\}$, and $\frac{1}{p_i}$ denotes the weight of sampling $x_t$. On this setting, the sampling process then adopts an error disagreement to control the hypothesis updates:
 \begin{equation}\label{eq:hypothesis-updates}
   \theta_{\rm \mathcal{A}}= \mathbb{E}_{x_t\in \mathcal{D}}   \mathop{{\rm sup}}\limits_{h\in B(h^*,r)} \left \{  \frac{ \ell(h(x_t),\mathcal{Y})-\ell(h^*(x_t),\mathcal{Y})   }{r}       \right\},
  \end{equation}
 where $\mathcal{D}$ denotes the margin distribution over  $\mathcal{X}$, drawing the candidate hypotheses. To reduce the complexity of the pruning process, one can shrink  $\mathcal{D}$  from the marginal distribution of $\mathcal{X}$, which derives the most of hypotheses, such as   \cite{dasgupta2008general,cohn1994improving}.
 
 Correspondingly,  $\ell(h(x), h'(x))$   denotes  the hypothesis disagreement  of $h$ and $h'$, which can be specified as the best-in-class error  on $\mathcal{Y}$
\begin{equation}\label{eq:loss}
\begin{split}
\ell(h(x),h'(x))= \Big|\mathop{ {\rm max}}\limits_{y \in \mathcal{Y}} \ell(h(x),y)-\ell(h'(x),y)\Big|,
\end{split}
\end{equation}
where   $y\in \mathcal{Y}$, and $\ell(h(x),h'(x))$ also can be simply written as $\ell(h,h')$.  Once the hypothesis update w.r.t. error  after adding $x_t$ is larger than $\theta_{\rm \mathcal{A}}$, the active learning algorithm $\mathcal{A}$  solicits $x_t$ as a significant update. Besides Eq.~(\ref{eq:loss}), $\ell(h(x),h'(x))$ also can be specified as all-in-class error \cite{cortes2020adaptive},  error entropy \cite{roy2001toward}, \emph{etc.} 

\subsection{Approximately Generalization Analysis}
From a hypothesis-pruning perspective, we firstly present a more general concept for   low-resource learning. Then, we present the generalization analysis of the convergence of the optimal hypothesis on error and label complexity bounds under a model-agnostic supervised and unsupervised fashion, respectively.
\subsubsection{A More General Concept }

A learning algorithm operates in a \emph{low-resource regime} if it attains a target generalization error $err(h)$ using only a  {limited number of labeled samples} $\mathcal{Q}$ drawn from the target distribution. Formally, for any hypothesis $h \in \mathcal{H}$ and any $\delta \in (0,1)$, with probability at least $1-\delta$, after $\mathcal{Q}$ samples, $err(h)$ converges to its optimal value and satisfies an upper bound \cite{dasgupta2008general}:  $err(h)+c \Big(\frac{1}{\mathcal{Q}}\big(d {\rm log}\mathcal{Q}+ {\rm log} \frac{1}{\delta}\big)+\sqrt{ \frac{err(h)}{\mathcal{Q}}{\mathcal{Q}(d {\rm log} \mathcal{Q}+ {\rm log}(\frac{1}{\delta}) }} \Big)$. By relaxing the constant $c$ and $err(h)$ ($err(h)<1$), 
the label complexity of any low-resource learning algorithm satisfies an upper bound of
\begin{equation}\label{eq:upper-bound}
N(\mathcal{H}, n, \mathcal{Q}, \mathcal{A}) \leq \mathcal{O}\Big(\frac{1}{\mathcal{Q}}\big(d {\rm log}\mathcal{Q}+ \rm{log} \frac{1}{\delta}\big) \Big).
\end{equation}
Eq.~(\ref{eq:upper-bound}) presents a coarse-grained observation on the upper bound of the label complexity. 
We next  introduce the error disagreement coefficient $\theta_\mathcal{A}$ to prune the hypothesis class.  
If the low-resource learning algorithm controls the hypothesis updates by Eq.~(\ref{eq:hypothesis-updates}), based on Theorem~2 of \cite{dasgupta2008general},  the expected label cost for the convergence of $err(h)$ is at most $1+c\theta_\mathcal{A}\Big(\big(d {\rm log} \mathcal{Q} + {\rm log} \frac{1}{\delta}\big) {\rm log} \mathcal{Q}\Big)$. By relaxing the constant $c$, we have 
\begin{equation}\label{eq:N}
N(\mathcal{H}, n, \mathcal{Q}, \mathcal{A}) \leq \mathcal{O}\Big( \theta_\mathcal{A}\big(d {\rm log} \mathcal{Q} + {\rm log} \frac{1}{\delta}\big) {\rm log} \mathcal{Q}\Big).
\end{equation}
With the inequalities of  Eqs.~(\ref{eq:upper-bound}) and (\ref{eq:N}), we present a more general concept for low-resource data.
\begin{insight}Properties of Hypotheses in Low-resource Learning.
With standard empirical risk minimization, learning low-resource data from $\mathcal{D}$ over $\mathcal{Q}$ times of sampling satisfies an incremental update on the optimal hypothesis with an error of  $err(h^*)$,
\begin{equation}
\begin{split}
& \argmin_{\mathcal{Q}} err(h_\mathcal{Q})\!\leq \!   \Bigg( err(h^*)+ \mathcal{O}\Big(\sqrt{err(h^*)  \Omega}  +\Omega  \Big) \Bigg), \ {\rm  s.t.} \ \  \Omega=  \frac{d   {\rm log} \mathcal{Q}+{\rm log}\frac{1}{\delta}}{ \mathcal{Q}}, \\
\end{split}
\end{equation}
where $h_\mathcal{Q}$ denotes the updated hypothesis at the $\mathcal{Q}$-time of sampling, yielding efficient generalization error  approximation  to the optimal $h^*$, that is, $h_\mathcal{Q}$ holds nearly consistent generalization ability  to $h^*$ in the hypothesis class $\mathcal{H}$.
\end{insight}

Insight~6 presents the hypothesis properties for learning with low-resource data which requires efficient generalization error approximation to that of  the optimal hypothesis relying on the original high-resource data representation. The insight is a model-agnostic setting via  $\mathcal{Q}$ times of i.i.d. sampling,   deriving an efficient approximation for the $\mathcal{Q}$-time hypothesis $h_\mathcal{Q}$ to the optimal $h^*$ in the hypothesis space $\mathcal{H}$.

\subsubsection{Learning with Low-resource Data}

With the  hypothesis preperties of learning from low-resource data, we next study that how to learn  low-resource data via empirical risk minimization (ERM), which can be generalized into different loss functions in real-world models. Our main theorem of the label complexity in regard to ERM is then presented in Theorems~\ref{Supervised-Fashion} and \ref{Unsupervised-Fashion}. 

Before presenting Theorem~\ref{Supervised-Fashion}, we need a technical lemma about the importance-weighted empirical risk minimization on $\ell(h_\mathcal{Q},h^*)$. The involved techniques  refer to the Corollary~4.2  of J. Langford et al.'s work  in \cite{langford2005tutorial},  and the  Theorem~1 of  C.~Sahyoun et al.'s work \cite{DBLP:conf/icml/BeygelzimerDL09}. 
 
\begin{lemma}
Let $R(h)$ be the   expected loss (also called learning risk) that stipulates   $R(h)=\mathbb{E}_{x\sim \mathcal{D}}[\ell(h(x),y)]$, and $R(h^*)$ be its  minimizer. On this setting, $\ell(h_\mathcal{Q},h^*)$ then can be bounded by  $\ell(h_\mathcal{Q},h^*) \leq R(h_\mathcal{Q})-R(h^*) $ that stipulates  $\mathcal{H}_{\mathcal{Q}}:=\{h\in \mathcal{H}_{\mathcal{Q}-1}: R(h_\mathcal{Q})\leq R(h^*)+2\Delta_{\mathcal{Q}-1} \}$, where $\Delta_{\mathcal{Q}-1}$ adopts a form \cite{cortes2020region} of 
\[ \frac{1}{\mathcal{Q}-1}  \Bigg[ \sqrt{ \Big[\sum_{s=1}^{\mathcal{Q}-1} p_s\Big]  \rm{log}\Big[\frac{(\mathcal{Q}-1)|\mathcal{H}|}{\delta}\Big]   }  +  \rm{log}\Big[\frac{(\mathcal{Q}-1)|\mathcal{H}|}{\delta}\Big]  \Bigg ], \]
where $|\mathcal{H}|$ denotes the number of hypothesis in  $\mathcal{H}$, and $\delta$ denotes a   probability threshold requiring $\delta>0$.  
Since $\sum_{s=1}^{\mathcal{Q}-1} p_s \leq \mathcal{Q}-1$, $\Delta_{\mathcal{Q}-1}$ can then be bounded by 
 \[\Delta_{\mathcal{Q}-1}=\sqrt{\Big(\frac{2}{\mathcal{Q}-1}\Big){\rm log}\Big(2\mathcal{Q}(\mathcal{Q}-1)\Big)  \frac{|\mathcal{H}|^2  }{\delta})},\] 
which denotes the loss disagreement bound to approximate a desired target hypothesis such that $R(h_\mathcal{Q})-R(h^*)\leq 2\Delta_{\mathcal{Q}-1}$.
\end{lemma}

There are two primary approaches to learning from low-resource data: supervised learning, which relies on a limited number of labeled examples, and unsupervised learning, which leverages unlabeled data to uncover inherent structure. A critical factor in supervised learning is label complexity, referring to the quantity of labeled samples required to achieve satisfactory performance. In contrast, unsupervised learning avoids the dependence on labels but often faces challenges in controlling generalization error due to the absence of explicit supervision. In the following sections, we present a detailed analysis of their respective generalization behaviors, examining how label complexity influences supervised methods and how unsupervised approaches manage generalization under scarce annotation conditions.

\emph{Supervised Fashion.}  \noindent We follow the setting of Lemma~1 to present the learning risk and label complexity for learning with low-resource data under $\mathcal{Q}$ rounds of importance sampling.

\begin{theorem}\label{Supervised-Fashion}
Given $\mathcal{Q}$ rounds of querying by employing the active learning algorithm $\mathcal{A}$,  with a probability at least  $1-\delta$, for all $\delta>0$, for any $\mathcal{Q}>0$,  the error disagreement of   $R(h_\mathcal{Q})$ and   $R(h^*)$ of learning with low-resource data is bounded by  
\begin{equation*}
\begin{split}
R(h_\mathcal{Q})-R(h^*) \leq  \max_{\mathcal{Q}} \Bigg\{\frac{2}{\mathcal{Q}} \Bigg[\sqrt{\sum_{t=1}^{\mathcal{Q}}p_t}+6\sqrt{{\rm log}\Big[\frac{2(3+\mathcal{Q})\mathcal{Q}^2}{\delta}\Big] } \Bigg] \times \sqrt{{\rm log}\Big[\frac{16\mathcal{Q}^2|\mathcal{H}_i|^2 {\rm log}\mathcal{Q}}{\delta}\Big]}\Bigg\}.
\end{split}
\end{equation*} 
Then, with a probability at least  $1-2\delta$, for all $\delta>0$, the label complexity of learning with low-resource data can be bounded by
\begin{equation*} 
\begin{split}
&N(\mathcal{H}, n, \mathcal{Q}, \mathcal{A}) \leq   \max_{ \mathcal{Q}} K_\ell\Bigg\{ \Big[\sum_{j=1}^{ \mathcal{Q}} \theta_{\mathcal{A}} R_j^* \mathcal{Q}  p_j\Big]   \!+\!\sum_{j=1}^{ \mathcal{Q}} O\Bigg(\sqrt{R_j^*\mathcal{Q}  p_j{\rm log}\Big[\frac{ \mathcal{Q}|\mathcal{H}_i| \mathcal{Q}}{\delta} \Big]}\Bigg)   \!  + \! O\Bigg( \mathcal{Q} {\rm log}^3\Big(\frac{\tau|\mathcal{H}_i| \mathcal{Q}}{\delta}\Big)\Bigg)\Bigg\},
\end{split}
\end{equation*} 
where $K_\ell$ is the slope asymmetry  over the   loss $\ell$,   $K_\ell= \mathop{{\rm sup}}\limits_{x_t', x_t\in \mathcal{D}} \left |\frac{{\rm max} \ \ell(h(x_t), \mathcal{Y})-\ell(h(x_t'), \mathcal{Y}) }    {{\rm min} \ \ell(h(x_t), \mathcal{Y})- \ell(h(x_t'), \mathcal{Y}) } \right|$, $R_j^*$ denotes the best-in-class risk at $j$-time querying, and $|\mathcal{H}|$ denotes the element number of  $\mathcal{H}$. 
\end{theorem}
The proofs of Theorem~1 and 2 of \cite{cortes2020region} can be adopted to prove the two inequalities of Theorem~\ref{Supervised-Fashion}, respectively.
 
 \emph{Unsupervised Fashion.} \noindent By employing unsupervised learning, the learning risk and label complexity of Theorem~\ref{Supervised-Fashion} are degenerated into a polynomial expression \cite{CaoMHE2022}. 

Given the input dataset $\mathcal{X}$ with $n$ samples, it is divided into $k$ clusters: $\{\mathcal{B}_1, \mathcal{B}_2,..., \mathcal{B}_k\}$, where  $\mathcal{B}_i$ has $N_i$ samples. Learning  low-resource data   performs IWAL for any $\mathcal{B}_i$. Specifically, it employs a new error disagreement $\theta_{\rm LSD}$ to control the hypothesis updates: 
\begin{equation}
 \theta_{\rm LLR}= \mathbb{E}_{x_t\in \mathcal{B}_i}   \mathop{{\rm sup}}\limits_{h\in B(h^*,r)} \left \{  \frac{ \ell(h(x_t),\mathcal{Y})-\ell(h^*(x_t),\mathcal{Y})   }{r}       \right\}.
\end{equation}
\begin{theorem}\label{Unsupervised-Fashion}
Given $T$ rounds of querying by employing the active learning algorithm $\mathcal{A}$, let $\mathcal{Q}$ be the number of ground-truth queries. If  learning low-resource data performs $\mathcal{A}$  for any $\mathcal{B}_i$, each cluster  will have $\tau=T/k$ rounds of querying. Then, with a probability at least  $1-\delta$, for all $\delta>0$, for any $\mathcal{Q}>0$,  the error disagreement of   $R(h_\tau)$ and   $R(h^*)$ of learning with low-resource data  is bounded by $k$ times of  polynomial 
\begin{equation*}
\begin{split}
 R(h_\tau)-R(h^*)    \leq k \times \max_{\mathcal{H}_i, i=1,2,...,k} \Bigg\{\frac{2}{\tau}  \Bigg[\sqrt{\sum_{t=1}^{\tau}p_t}+6\sqrt{{\rm log}\Big[\frac{2(3+\tau)\tau^2}{\delta}\Big] } \Bigg] \times \sqrt{{\rm log}\Big[\frac{16\tau^2|\mathcal{H}_i|^2 {\rm log}\tau}{\delta}\Big]}\Bigg\},
\end{split}
\end{equation*} 
Then, with a probability at least  $1-2\delta$, for all $\delta>0$, the label complexity of learning with low-resource data can be bounded by
\begin{equation*} 
\begin{split}
&N(\mathcal{H}, n, \mathcal{Q}, \mathcal{A})  \leq 8k \times \max_{\mathcal{H}_i, i=1,2,...,k} K_\ell\Bigg\{ \Big[\sum_{j=1}^{N_i} \theta_{\rm LLR} R_j^*\tau p_j\Big] 
   +\sum_{j=1}^{N_i} O\Bigg(\sqrt{R_j^*\tau p_j{\rm log}\Big[\frac{\tau|\mathcal{H}_i|N_i}{\delta} \Big]}\Bigg)   \!+\! O\Bigg(N_i {\rm log}^3\Big(\frac{\tau|\mathcal{H}_i|N_i}{\delta}\Big)\Bigg)\Bigg\}.
\end{split}
\end{equation*} 
where $K_\ell$ is the slope asymmetry over the limited loss $\ell$ on $\mathcal{B}_i$, \textit{i.e.}, 
$\ell_{\mathcal{B}_i}$,   $K_\ell= \mathop{{\rm sup}}\limits_{x_t', x_t\in \mathcal{B}_i} \left |\frac{{\rm max} \ \ell_{\mathcal{B}_i}(h(x_t), \mathcal{Y})-\ell_{\mathcal{B}_i}(h(x_t'), \mathcal{Y}) }    {{\rm min} \ \ell_{\mathcal{B}_i}(h(x_t), \mathcal{Y})- \ell_{\mathcal{B}_i}(h(x_t'), \mathcal{Y}) } \right|$, $R_j^*$ denotes the best-in-class risk at $j$-time querying, and $|\mathcal{H}|$ denotes the element number of  $\mathcal{H}$. More details and proofs are presented in  Supplementary Material.
\end{theorem}

In summary, learning from low-resource data can approximately converge to the optimal hypothesis $h^*$ in either a supervised or unsupervised fashion, minimizing the disagreement to its risk $R(h^*)$ with bounded loss, while achieving reduced label complexity, and acceptable approximation error. This analysis forms the foundation for our survey, guiding the investigation of optimization methods, the organization of scenarios, and the identification of key challenges in the field.


\section{Investigation from Optimization}

The theoretical analysis provides effective guarantees for learning from low-resource data through controlled bounds. Building on this foundation, we initiate our exploration of potential optimization strategies tailored to low-resource scenarios, including gradient-informed optimization, meta-iteration optimization, geometry-aware optimization, and LLMs-powered optimization.

\subsection{Gradient-Informed Optimization}
To explore optimization solvers for the aforementioned geometric representations, we categorize potential methods into three gradient-based approaches: (1) Euclidean gradient descent, suitable for optimizing Euclidean geometric representations; (2) Riemannian gradient descent, tailored for hyperbolic geometric representations; and (3) Stein variational gradient descent (SVGD), which can be applied to both Euclidean and hyperbolic spaces. Notably, SVGD has been extended to Riemannian manifolds, enabling its application in non-Euclidean geometries . These optimization methods are also effective in addressing challenges associated with high-resource data scenarios, such as enhancing computational efficiency and scalability .
\subsubsection{Euclidean Gradients}
Stochastic Gradient Descent (SGD) \cite{ruder2016overview} is an effective approach to find the local minima of a cost function, it can be adopted to optimize Euclidean centroids which are formulated as an argmin problem in Euclidean space. \\
\textbf{Stochastic gradient descent.} Given a minimization problem of $\min \limits_{x\in \mathbb{R}^{n} } J(x)$ in Euclidean space, at $t$-time, the parameters $x_t$ is updated as \cite{ruder2016overview}:
\begin{equation}
\begin{split}
x_{t + 1} = x_t - \eta \cdot \nabla_{x}J(x),
\end{split}  
\end{equation}
where $J(x)$ denotes the cost function parameterized by $x$, and $\eta$ denotes the learning rate.
\subsubsection{Non-Euclidean Gradients}
Manifold optimization \cite{boumal2020introduction} aims to find solutions for various constrained optimization problems in Euclidean space by transforming them into unconstrained optimization problems on Riemannian manifolds. To perform iterative optimization on these manifolds, Riemannian Gradient Descent (RGD) \cite{chen2021decentralized,liu2018riemannian} is introduced. This approach has led to rapid development in Riemannian optimization. Interestingly, RGD is also employed in hyperbolic geometry to optimize different paradigms on the Poincaré ball $\mathcal{P}^{n}$ and the Lorentz model $\mathcal{L}^{n}$.\\
 \textbf{Riemannian gradient descent.} Given a minimization problem of $\min \limits_{x \in \mathcal{M}} J(x)$ on a Riemannian  manifold $\mathcal{M}$,   $x_t$ at $t$-time  is updated by the  exponential map  ${\rm exp}_x$  \cite{chen2021decentralized}: 
\begin{equation}
\begin{split}
 x_{t+1}={\rm exp}_{x_t}\Big(-\eta J'(x_t)\Big),
\end{split}
\end{equation}
 where $J'(x_t)$ denotes the Riemannian gradient on the tangent space $\mathcal{T}_x \mathcal{M}$ and $\eta$ denotes the learning rate. \\
\textbf{Exponential map on Poincaré model.} Given a Riemannian metric $g_{x}(\cdot,\cdot)$ that induces an inner product $\langle u,v\rangle:=g_x(u,v)$ on tangent space $\mathcal{T}_x\mathcal{M}$. For each point $x\in\mathcal{M}$ and vector $u\in \mathcal{T}_x\mathcal{M}$, there exists a unique geodesic $\gamma: [0,1]\rightarrow \mathcal{M}$ where $\gamma(0)=x,\gamma'=u$.
The exponential map ${\rm exp}_{x} : \mathcal{T}_x\mathcal{M} \rightarrow \mathcal{M}$ is defined as ${\rm exp}_{x}(u)=\gamma(1)$, where $d_\mathcal{P}(x, {\rm exp}_{x}(u))=\sqrt{g_\mathcal{M}(u,u)}$. With  \cite{lou2020differentiating},
  \begin{equation}
\begin{split}
{\rm exp}_{x}(u)=\frac{(1-2\langle x,z\rangle_2-\|z\|^2)x+(1+\|x\|^2)z}{1-2\langle x,z\rangle_2+\|x\|^2\|z\|^2}, 
\end{split}
\end{equation}
 where $z={\rm tanh}( \frac{\|u\|^2}{1+\|x\|^2})\frac{u}{\|u\|}$. \\
\textbf{Exponential map on Lorentz model.} With Proposition~3.2 of  \cite{chami2019hyperbolic}, ${\rm exp}_{x}(u)=\gamma(1)$ on a Lorentz model $\mathcal{L}^{n}$ is expressed as  
 \begin{equation}
\begin{split}
{\rm exp}_{x}(u)={ \rm cosh}(\|u\|_\mathcal{L})x+u\frac{{\rm sinh}(\|u\|_\mathcal{L})}{\|u\|_\mathcal{L}}.
\end{split}
\end{equation}

\subsubsection{Stein Gradient}
Bayesian inference \cite{box2011bayesian} is a statistical inference that invokes the Bayes theorem to approximate the probability distribution. Variational inference \cite{blei2017variational} approximates parameterized distribution through  probabilistic optimization that involves sampling tractable variables, such as Markov Chain Monte-Carlo (MCMC). However, the approximation errors of both bayesian and variational inference on estimation over likelihoods or posterior parameter distribution are not easy to control, which can result in unstatistically significant results with calibration. To effectively tight the approximation, Liu et al. \cite{chwialkowski2016kernel} adopt the Stein operation which controls the bounds on the distance between two probability distributions in a given probability metric. With this proposal, Liu et al. then propose the Stein Variational Gradient Descent (SVGD) algorithm \cite{liu2019stein} that minimizes the KL divergence \cite{hershey2007approximating} of two probability distributions $p$ and $q$ by utilizing Kernelized Stein Discrepancy (KSD) and smooth transforms, thus conducting iterative probability distribution approximation. 
\par In detail, MCMC estimates the denominator integral of posterior distribution through sampling, which results in the issue of computational inefficiencies. Let $p_0(x)$ be the prior, $\{D_i\}_{i=1}^{N}$ be a set of i.i.d. observations, and $\Omega=\{q(x)\}$ be the distribution set, variational inference adopts a novel idea to alleviate the issue by minimizing the KL Divergence between the target posterior distribution $p(x)$ and another distribution $q^*(x)$ so as to approximate $p(x)$:
\begin{equation}\label{eq:vi-kl}
\begin{split}
{{q^*(x)}} & = \argmin \limits_{q(x) \in \Omega} \Big\{{\rm KL}(q(x) || p(x))\equiv \mathcal{W}\Big\},
\end{split}
\end{equation}
where $\overline p(x) := p_0(x) \prod \limits_{i=1}^N p(D_i | x)$, ${Z= \int {\overline p(x)}\dif x}$ denotes the normalization constant which requires complex calculations, and $\mathcal{W} = \mathbb{E}_q [\log{q(x)}] - \mathbb{E}_q [{\rm log} {\overline p(x)}] + \log Z$. Hence, to circumvent the computation of troublesome normalization constant $Z$ and seek for a general purpose bayesian inference algorithm, Liu et al. adopt the Stein methods and propose the SVGD algorithm. More details are presented in Supplementary Material.
\par Given notions of Stein’s Identity in Eq.~(\ref{eq:SI}), Stein Discrepancy in Eq.~(\ref{eq:SD}) and Kernelized Stein Discrepancy in Eq.~(\ref{eq:KSD}) of the Stein methods, Liu et al. rethink the goal of variational inference which is defined in Eq.~(\ref{eq:vi-kl}), they consider the distribution set $\Omega$ could be obtained by smooth transforms from a tractable reference distribution $q_0(x)$ where $\Omega$ denotes the set of distributions of random variables which takes the form $r = T(x)$ with density:
\begin{equation}\label{eq:density1}
\begin{split}
q_{[T]}(r) = q(\mathcal{R}) \cdot |\det(\nabla_r \mathcal{R})|,
\end{split}
\end{equation}
where $T: \mathcal{X} \to \mathcal{X}$ denotes a smooth transform,  $\mathcal{R}=T^{-1}(r)$ denotes the inverse map of $T(r)$ and $\nabla_r \mathcal{R}$ denotes the Jacobian matrix of $\mathcal{R}$. With the density, there should exist some restrictions for $T$ to ensure the variational optimization in Eq.~(\ref{eq:vi-kl}) feasible. For instance, $T$ must be a one-to-one transform, its corresponding Jacobian matrix should not be computationally intractable. Also, with \cite{rezende2015variational}, it is hard to screen out the optimal parameters for $T$.
\par Therefore, to bypass the above restrictions and minimize the KL divergence in Eq.~(\ref{eq:vi-kl}), an incremental transform ${T(x) = x + \varepsilon \varphi(x)}$ is proposed, where $\varphi(x)$ denotes the smooth function controlling the perturbation direction and $\varepsilon$ denotes the perturbation magnitude. With the knowledge of Theorem \ref{Stein-KL} and Lemma \ref{Stein-varphi}, how can we approximate the target distribution $p$ from an initial reference distribution $q_0$ in finite transforms with  $T(x)$? Let $s$ denote the total distribution number, an iterative procedure which can obtain a path of distributions $\{q_t\}^s_{t=1}$ via Eq.~(\ref{eq:iterative-procedure}) is adopted to answer this question:
\begin{equation}\label{eq:iterative-procedure}
\begin{split}
& \quad \quad q_{t+1} = q_t[T^*_t],\\
& T^*_t(x) = x + \varepsilon_t \varphi^*_{q_t, p}(x),
\end{split}
\end{equation}
where $T^*_t$ denotes the transform direction at iteration $t$, which then decreases the KL Divergence with $\varepsilon_t {{\rm KSD}(q_t, p)}$ at {$t$-th} iteration. Then, the distribution $q_t$ finally converges into the target distribution $p$.
To perform above iterative procedure, Stein Variational Gradient Descent (SVGD) adopts the iterative update procedure for particles which is presented in Theorem \ref{Stein-iteration} to approximate $\varphi^*_{q, p}$ in Eq.~(\ref{eq:varphi}).
\begin{theorem}\label{Stein-iteration}
Let $p(x)$ denote the target distribution, $\{x^0_i\}^m_{i=1}$ denote the initial particles \cite{liu2019stein}. 
Also, at iteration $t$, let ${\vartheta} = \nabla_{x^t_j} {\log} p(x^t_j)$, ${\mu}(x^t_j, x)=\nabla_{x^t_j} k(x^t_j, x)$ denote a regular term, $\Phi$ denote $\varepsilon_t \hat{\varphi}^*(x)$, the particles set are updated iteratively with $T^*_t$ defined in Eq.~(\ref{eq:iterative-procedure}), in which the expectation under $q_t$ in $\varphi^*_{q_t, p}$
is approximated by the empirical mean of $\{x^t_i\}^m_{i=1}$:
\begin{equation}
x^{t}_i + \Phi \rightarrow x^{t + 1}_i,
\end{equation}
where 
\begin{equation}
\hat{\varphi}^*(x) = \frac{1}{m} \sum_{j=1}^{m} [k(x^t_j, x) {\vartheta} + {\mu(x^t_j, x)}].
\end{equation}
\end{theorem}
Regarding $\hat{\varphi}^*(x)$, the first term $k(x^t_j, x) {\vartheta}$ denotes the weighted sum of the gradients of all the points weighted by the kernel function, which follows a smooth gradient direction to drive the particles towards the probability areas of $p(x)$; The second term $\mu(x^t_j, x)$ denotes a regular term to prevent the collapse of points into local modes of $p(x)$, \textit{i.e.}, pushing $x$ away from $x^t_j$.

\subsection{Meta-Iteration Optimization}


Theoretically, learning parameters could be optimized by 
utilizing the well-generalized meta-knowledge across various learning tasks, namely, teach the \emph{learning} models \emph{to learn} for unseen tasks \cite{hospedales2020metalearning}. This requires that the current parameter update policy could be effectively generalized into unseen tasks. Therefore, \emph{how to extract well-generalized meta-knowledge} \cite{tian2021consistent} has become an critical issue in meta parameter update. To explore potential meta parameter update fashions, we integrate meta-learning from multiple fashions including bilevel fashion, task-distribution fashion, and feed-forward fashion as follows. 


\subsubsection{Bilevel Fashion}
\begin{figure}
    \centering
    \includegraphics[width=0.95\textwidth]{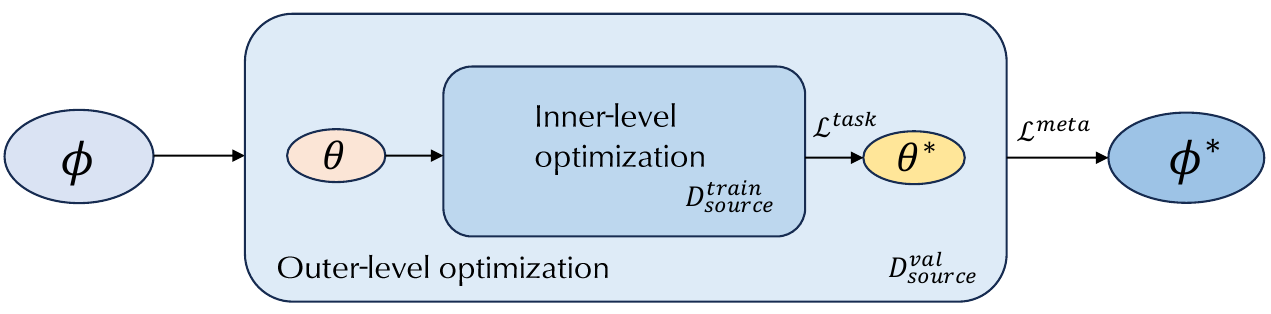}
    \caption{The bilevel optimization fashion of meta-learning. During the meta-training phrase, the inner-level optimization is to train the model on different tasks which are sampled from $D^{train}_{source}$ to obtain the optimal model with parameters ${\theta}^{\ast}$, $\mathcal{L}^{task}$ denotes the optimization objective for inner-level optimization; the outer-level optimization aims to obtain a general meta-knowledge ${\phi}^{\ast}$ which be quickly adapted to unseen tasks. $\mathcal{L}^{meta}$ denotes the optimization objective to obtain ${\phi}^{\ast}$.}
    \Description{The bilevel optimization fashion of meta-learning.}
    \label{fig:meta-learning}
    \vspace{-0.3cm}
\end{figure}

Bilevel optimization \cite{ji2021bilevel,khanduri2021near,yang2021provably} is a hierarchical optimization problem, which means one optimization objective contains another inner optimization objective as a constraint. The bilevel optimization scheme of meta-learning is presented in Figure \ref{fig:meta-learning}. Therefore, from the perspective of this \cite{rajeswaran2019meta,wang2020global}, the meta-training phrase is formalized as:
\begin{equation}
\begin{split}
&\phi^{\ast} = \argmin \limits_{\phi} \sum_{i=1}^{M}{\mathcal{L}^{meta}{\Big(}{\theta^{\ast}}^{(i)}(\phi)}, \phi, {D_{source}^{val}}^{(i)}\Big) \\
&{\rm s.t.} \quad {{{\theta^{\ast}}^{(i)}(\phi)} = \argmin \limits_
{\phi} {\mathcal{L}^{task}\Big(\theta, \phi, {D_{source}^{train}}^{(i)}\Big)}},
\end{split}
\end{equation}
where $\mathcal{L}^{meta}$ and $\mathcal{L}^{task}$ denote the outer and the inner loss objectives, respectively. The inner-level optimization is with $\phi$ which is defined in outer-level optimization as condition, but $\phi$ cannot be changed during the inner-level optimization phrase; the outer-level optimization utilize ${{\theta^{\ast}}^{(i)}(\phi)}$ obtained from the inner-level optimization to optimize the meta-knowledge $\phi$. 

\subsubsection{Task-Distribution Fashion}
From the perspective of task-distribution \cite{vuorio2019multimodal,collins2020task}, meta parameter update considers tasks as samples of the model. Besides, this update paradigm aims to learn a common learning algorithm which can generalize across tasks. In detail, the goal is to learn a general meta-knowledge $\phi$ which can minimize the expected loss of meta-tasks. Let $q(\mathcal{T})$ denote the distribution of tasks, $D$ denote the dataset for meta-tasks,  meta-task optimization (meta-learning) can be formalized as:
\begin{equation}
\begin{split}
\min \limits_{\phi} \underset{\mathcal{T} \sim q(\mathcal{T})}{\mathbb{E}} \mathcal{L}(D; \phi),
\end{split}
\end{equation}
where $\mathcal{L}(D; \phi)$ denotes the loss function to measure the performance of  the learning model.
To address the above optimization problem, it is assumed that we can obtain a set of source tasks sampled from $q(\mathcal{T})$.
With \cite{hospedales2020metalearning},  the above optimization consists of two phases: \textbf{meta-training} phrase and \textbf{meta-testing} phrase.
Let $\mathcal{D}_{source} = \{(D_{source}^{train}, D_{source}^{val})^{(i)}\}_{i=1}^{M}$ denote the set of $M$ source tasks in meta-training phase, where $D_{source}^{train}$ and $D_{source}^{val}$ denote the training and validation data respectively for source tasks, the serial number $i$ indicates each task; $\mathcal{D}_{target} = \{(D_{target}^{train}, D_{target}^{test})^{(i)}\}_{i=1}^N$ denote the set of $N$ target tasks in meta-testing phase, where $D_{target}^{train}$ and $D_{target}^{test}$ denote the training and testing data for target tasks,  respectively. On this setting, the meta-training phrase is to learn the optimal meta-knowledge $\phi^{\ast}$ and maximize the log likelihood by sampling different source tasks from $\mathcal{D}_{source}$, it is thus formalized as:
\begin{equation}
\begin{split}
\max \limits_{\phi} \log {p(\phi|\mathcal{D}_{source})}.
\end{split}
\end{equation}
By solving the maximum problem, we obtain a well-generalized meta-knowledge $\phi^{\ast}$, which is utilized to facilitate the model learning on unseen target tasks. The meta-testing phase aims to obtain a robust model on the training data of each unseen target task sampled from $\mathcal{D}_{target}$ with the help of $\phi^{\ast}$, which can be formalized as:
\begin{equation}
\begin{split}
\max \limits_{\theta} \log {p\Big(\theta|\phi^{\ast}, {{D_{target}^{train}}\Big)}}.
\end{split}
\end{equation}
We can thus obtain a model with parameters ${\theta}^{\ast}$ by solving the above maximum problem and evaluate its performance by conduct target tasks which are sampled from $D_{target}^{test}$.

\subsubsection{Feed-Forward Fashion}
With \cite{hospedales2020metalearning}, there exists different meta-learning methods which synthesize models in feed-forward manners \cite{bertinetto2016learning,denevi2018learning}. Let $\gamma = \mathbf{x}^{\top} \mathbf{e}_{\phi}\left(\mathcal{D}^{train}\right)$, a simple example for meta-training linear regression objective which optimizes over a distribution of meta-training tasks from feed-forward perspective is defined as:
\begin{equation}
\min _{\phi} \underset{\mathcal{T} \sim q(\mathcal{T})}{\mathbb{E}} \sum_{(\mathbf{x}, y) \in \mathcal{D}^{val}}\left[\left(\gamma -y\right)^{2}\right],
\end{equation}
where the training set $\mathcal{D}^{train}$ is embedded into the vector $\mathbf{e}_{\phi}$ which defines the linear regression weights, thus making prediction of samples $\mathbf{x}$ from the validation set $\mathcal{D}^{val}$.

It is noteworthy that meta-learning could perform well on optimizing low-resource data training. Namely, we could provide a potential parameter optimization over low-resource data training with the power of meta-learning. In bilevel fashion, during the inner-level optimization phrase, the parameters could be optimized over low-resource data training tasks; and the meta-knowledge could be efficiently updated during the outer-level optimization phrase. In task-distribution fashion, meta-learning could efficiently guide the model update in meta-training phrase under low-resource data scenarios, and generalize well across unseen task distributions in meta-testing phrase. In feed-forward fashion, meta-learning could promote the parameter update over training task distributions in feed-forward manner under low-resource data setting. Based on above fashions, the well-generalized meta-knowledge could be efficiently extracted. It's promising to introduce the perspective of meta parameter optimization over low-resource data training, and awaits futher exploration.

\subsection{Other Optimization Methods}

\paragraph{Geometry-Aware Optimization.} Gradients and their parameter updates usually remain powerful and versatile optimization methods in Euclidean space. Geometry-aware optimization techniques offer specialized advantages in contexts where the data or model structure aligns with non-Euclidean geometries, such as manifold covers, hierarchical structures, graphs, and trees. In these cases, the mean update is a typical approach for optimizing geometric representations. Therefore, we elaborate from four dimensions: Fréchet Mean, Euclidean Mean, Non-Euclidean Mean, and Kernel Mean. For the specific optimization details, please refer to A1 part in the Supplementary Material.

\paragraph{LLMs-Powered Optimization.} Optimization problems are core challenges across numerous fields, while traditional methods often rely on extensive data, computational resources, or human intervention. In recent years, LLMs with their natural language understanding and generation capabilities, have provided a new perspective for optimization tasks, particularly demonstrating unique advantages in low-resource, low-human-involvement, and few-shot scenarios. This section reviews the latest advancements of LLMs in the optimization domain, categorized into four aspects: gradient optimization, prompt optimization, evolutionary strategy generation optimization, and scientific problem optimization. For detailed investigation, please refer to A2 part in the Supplementary Material.

\section{Investigation from Scenario}

The optimization strategies outlined above offer promising pathways for learning from low-resource data. As interest in this area continues to grow, the research community is actively exploring ways to enhance model training by enabling more efficient representations of limited data across a range of learning paradigms. Notably, in practical scenarios, a growing body of research has focused on enabling agents to function autonomously in low-resource environments, making independent decisions to complete tasks. This trend draws on large models for knowledge transfer and generalization, with hierarchical reinforcement learning providing mechanisms to decompose complex problems into tractable sub-tasks.

\begin{figure} 
  \centering  
  \includegraphics[width=0.95\textwidth]{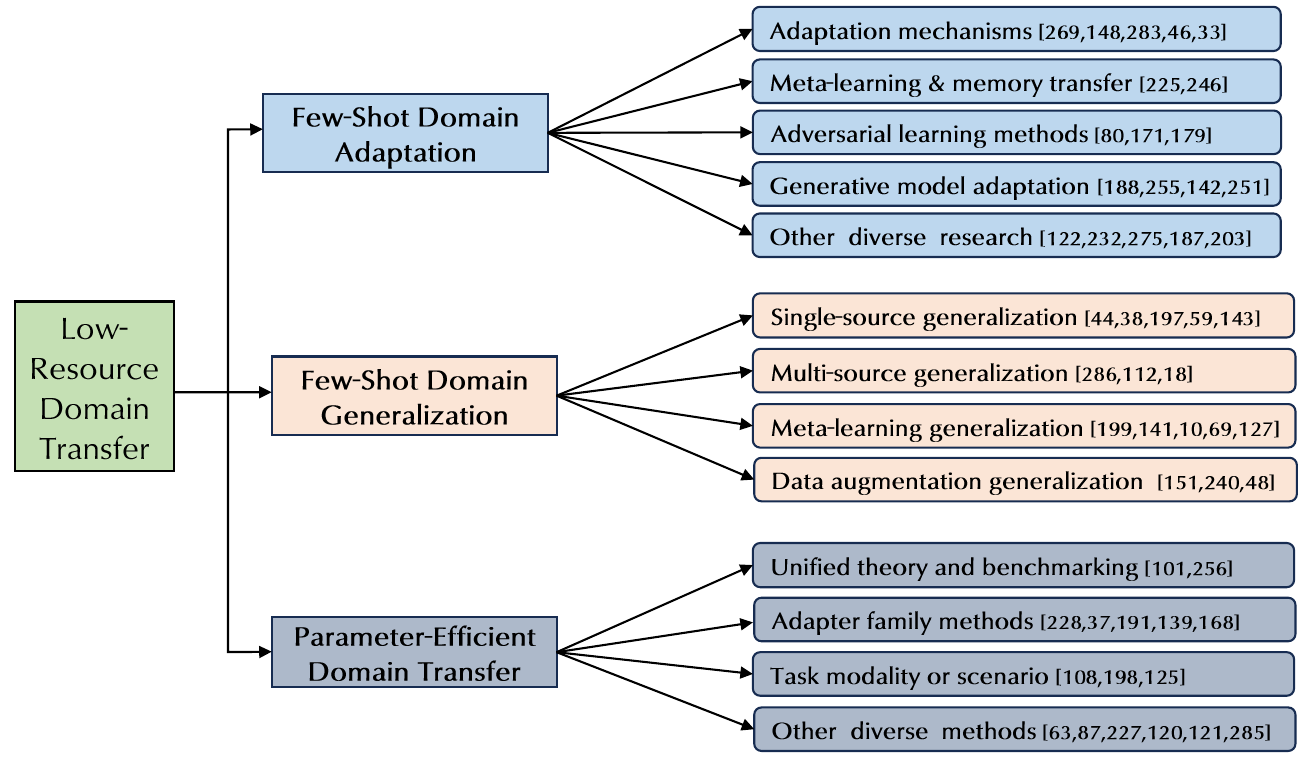}  
  \caption{\cao{Framework of efficient exploration for low-resource domain transfer.}}  
  \Description{The framework illustrates the process of efficient exploration for low-resource domain transfer.}
  \label{fig:Low-Resource Investigation} 
  \vspace{-0.2cm}
\end{figure}

These foundations collectively support agents in transferring knowledge, generalizing under resource constraints, and modeling tasks hierarchically. Furthermore, expanding these capabilities through swarm intelligence approaches based on hierarchical network collaboration brings us closer to artificial general intelligence. Thus, in the following part, we focus on three key areas: domain transfer and generalization, reinforcement learning, and hierarchical structure modeling in low-resource scenarios. The organization of their content is shown in Figure \ref{fig:Low-Resource Investigation}.

\subsection{Domain Transfer with Low-resource Data}

\par The core value of low-resource domain transfer lies in overcoming the limitation of training each task from scratch by leveraging knowledge from data-rich source scenarios to address the challenges of target tasks under \textbf{data scarcity}, \textbf{limited computing resources}, and \textbf{distribution shift or unknown domains}. This paradigm enables efficient and generalizable learning in resource-constrained environments. The methods of low-resource domain transfer can be roughly divided into three key areas: 1) \textbf{Few-Shot Domain Adaptation}, which focuses on rapid adaptation to new tasks with limited samples through techniques like embedding adaptation and meta-learning; 2) \textbf{Few-Shot Domain Generalization}, which tackles generalization to unknown target domains with scarce labeled data; and 3) \textbf{Parameter-Efficient Domain Transfer}, which optimizes resource usage by fine-tuning only a subset of model parameters. These approaches collectively address the diverse challenges of low-resource scenarios, as detailed in the following sections.

\textbf{\underline{ \textcircled{1} Few-Shot Domain Adaptation.}}  This theme aims to quickly adapt to new tasks using limited samples, and often integrates methods such as embedding adaptation, meta-learning, pre-training, knowledge transfer, and generative adversarial learning. Based on core technical ideas, it can be categorized as follows:
\begin{enumerate}[label=\arabic*)]
    \item \textbf{Adaptation mechanisms}, which include embedding layer adaptation \cite{ye2020fewshot}, task-specific adapters \cite{9879070}, pre-trained network adaptation \cite{zhang2022lccs}, one-step hypothesis adaptation \cite{chi2021tohan}, and optimization of domain-specific component Batch Normalization \cite{8953938}, focus on narrowing domain gaps through module or parameter adjustments.
    \item \textbf{Meta-learning \& memory transfer}, which covers meta-transfer learning \cite{8954051} and meta-memory transfer \cite{10224305}, leverages the meta-learning paradigm to quickly adapt to new domain tasks.
    \item \textbf{Adversarial learning-driven methods}, such as \cite{NIPS2017_21c5bba1, Luo2020ASM, 2946704}, achieve domain feature alignment through adversarial training.
    \item \textbf{Generative model adaptation}, which focuses on cross-domain image generation \cite{9577580}, generative model structure alignment \cite{9879663}, hybrid adaptation \cite{ICLR2024_dbb8193a}, and Re-modulation adaptation \cite{NEURIPS2023_b2e20d74}, uses generative methods to supplement information in low-resource few-shot domains.
    \item \textbf{Other diverse research}, which includes causality and data augmentation \cite{Teshima2020Fewshot,10098552}, prototype self-supervision for unsupervised domain adaptation \cite{9577429}, theoretical analysis and cross-domain characteristic understanding \cite{oh2022understanding}, and adaptation for specific application scenarios \cite{raghuram2023fewshot}, covers diverse research directions.
\end{enumerate}

\textbf{\underline{ \textcircled{2} Few-Shot Domain Generalization.}}
Compared with domain adaptation, which leverages partial target domain data to assist in adaptation, the key characteristic of domain generalization is that the target domain distribution remains fully unknown during the training phase. As a scenario more aligned with real-world low-resource settings, few-shot domain generalization further requires addressing the challenge of scarce labeled samples in the source domain. Focusing on this low-resource few-shot domain generalization problem, it can be roughly categorized into the following types by core technical paths and task scenarios:
\begin{enumerate}[label=\arabic*)]
    \item \textbf{Single-source domain generalization}, which focuses on scenarios involving only one source domain and includes the generalization of counting tasks based on universal representation matching \cite{Chen_2025_CVPR}, causal meta-learning \cite{Chen_2023_CVPR}, single-domain generalization meta-learning \cite{9157002}, single-source cross-domain few-shot frameworks \cite{das2022confess}, and progressive domain expansion networks \cite{9578057}.
    \item \textbf{Multi-source domain generalization}, which designs strategies for multi-source domain settings and includes progressive mix-up multi-source domain transfer \cite{zhu2023progressive}, switch learning \cite{hu2022switch}, and marginal transfer learning \cite{3546260}.
    \item \textbf{Meta-learning generalization},  related research includes two-level meta-learning \cite{10203333}, basic meta-learning frameworks \cite{Li_Yang_Song_Hospedales_2018}, meta-regularization \cite{3327036}, cross-domain few-shot batch normalization \cite{du2021metanorm}, and review studies in this field \cite{Khoee2024DomainGT}, and its core lies in enabling models to quickly acquire cross-domain adaptation capabilities through the meta-learning paradigm.
    \item \textbf{ Data augmentation generalization},  which improves generalization by enhancing data diversity or optimizing feature representations, with examples including noise-enhanced supervised autoencoders \cite{9709997}, adversarial data augmentation \cite{3327439}, zero-shot dataset generation \cite{emnlp_ChoiKYYK24}.
\end{enumerate}

\textbf{\underline{ \textcircled{3} Parameter-Efficient Domain Transfer.}}
Traditional transfer learning often requires full parameter fine-tuning of pre-trained models. This not only incurs substantial computing and storage resource consumption but also is prone to causing overfitting in few-shot or low-resource scenarios. In contrast, Parameter-Efficient Transfer Learning (PETL) achieves efficient transfer by adjusting only a subset of key model parameters, such as inserting lightweight adapters or optimizing specific modules. This approach significantly reduces resource consumption while maintaining generalization performance. Studies on PETL can be roughly categorized into the following types:
\begin{enumerate}[label=\arabic*)]
    \item \textbf{Unified theory and benchmarking}, which focuses on the common laws and evaluation systems of PETL, includes studies that construct a general theoretical perspective \cite{he2022towards} and work that builds dedicated benchmarks for visual PETL \cite{NEURIPS2024_935de67d}, providing a unified analytical framework and evaluation standards for the field.
    \item \textbf{Adapter family methods}, which covers Adapter variants adapted to different modalities and tasks, such as vision-language task adaptation \cite{9878858}, convolutional network adaptation \cite{Chen_2024_CVPR}, image-to-video transfer adaptation \cite{3602189}, conditional adaptation with fast inference capability \cite{NEURIPS2023_19d7204a}, and cross-modal unified adaptation \cite{ICLR2024_6b055b95}.
    \item \textbf{Task modality or scenario}, which includes studies exclusive to the NLP field \cite{pmlr-v97-houlsby19a}, visual-language navigation tasks \cite{10378091}, and contrastive alignment from vision to language \cite{khan2023contrastive}, aims to address the parameter-efficient transfer needs of different domains in a targeted manner.
    \item \textbf{Other diverse methods}, which include designing high-accuracy and efficient memory \cite{SHERL24}, pruning \cite{guo-etal-2021-parameter}, ladder side-tuning \cite{3601214}, redundancy-aware tuning \cite{10204861}, low-rank decomposition and alignment \cite{Jiang2022}, and the combination of Adapter and prompt tuning \cite{10655892}, and this category further expands the technical implementation paths of PETL.
\end{enumerate}

\underline{ \textcircled{4}   \textbf{Advances.}}  
Adapting pre-trained models with small, task-specific datasets has proven to be more efficient and effective than training from scratch, offering both parameter efficiency and improved robustness. By selectively fine-tuning only the most relevant parameters, such approaches reduce annotation costs, accelerate training, and enhance generalization even under sparse or noisy data conditions. These advantages are especially critical in domain adaptation and domain generalization, where models must maintain performance across distribution shifts and unseen domains. Low-resource techniques have been explored across transfer learning strategies: instance-based methods that re-weight informative source samples \cite{xia2013instance}; feature-based methods that extract domain-invariant representations \cite{gong2013connecting, li2018domain, phung2021learning}; and network-based methods that reuse pre-trained networks or transferable representations \cite{zhuang2015supervised}.

With the advent of foundation models, these strategies have evolved further. Parameter-efficient tuning methods, such as prompt tuning\cite{lester2021power} and LoRA\cite{hu2022lora}, facilitate domain adaptation without the need for full retraining. Meanwhile, knowledge distillation from large models condenses general knowledge into compact, task-specific learners. Such approaches improve cross-domain generalization while minimizing data, annotation, and parameter demands \cite{raghu2019transfusion}. Nevertheless, knowledge transfer can suffer from negative transfer when source and target domains misalign \cite{jiang2022transferability}. Low-resource learning mitigates this risk by filtering detrimental knowledge and guiding models toward compact, task-focused representations. Despite promising progress, seamlessly integrating these large-model–driven, low-resource techniques into standard pipelines for domain adaptation and domain generalization remains an open research challenge.

\subsection{Reinforcement Feedback with Low-resource Data}

\begin{figure}
    \centering
    \includegraphics[width=0.95\textwidth]{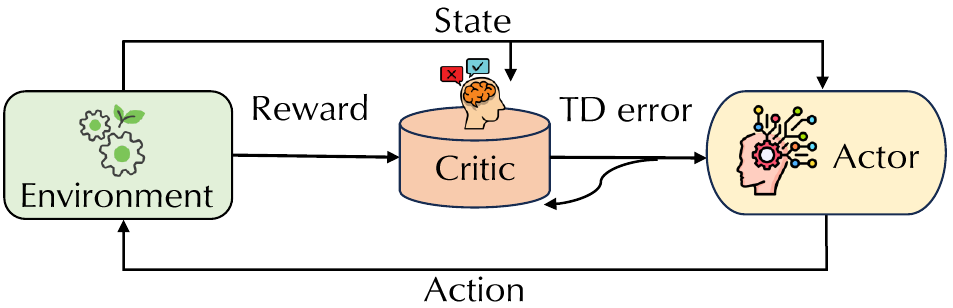}
    \caption{Reinforcement learning actor-critic strategies involve three components: Environment, Critic (value function), and Actor (policy). The Environment provides states and rewards based on the Actor’s actions; the Actor selects actions from states, while the Critic evaluates state or state–action values. TD error measures the gap between predicted and actual returns, driving updates for both the Critic and the Actor.}
    \Description{actor-critic strategies}
    \label{fig:RL-AC}
\end{figure}

\begin{table}[htbp]
\centering
\caption{\cao{Classification and Statistics of Research Literature on Low-Resource  Reinforcement Learning.}}
\begin{tabular}{
  >{\arraybackslash}p{2.5cm} 
  >{\arraybackslash}p{6.0cm} 
  >{\arraybackslash}p{5.0cm}   
} 
\hline
\textbf{Research} & \textbf{Subcategory} & \textbf{Representative Papers} \\ 
\hline
\multirow{4}{=}{\centering Sample Efficient RL} 
& Model-based methods
& \cite{122377}, \cite{hafner2019learning}, \cite{NEURIPS2019_5faf461e}, \cite{NEURIPS2018_f02208a0}, \cite{3327385}, \cite{Hafner2020Dream},  \cite{hafner2021mastering}, \cite{NEURIPS2021_d5eca8dc}, \cite{curi2020efficient}, 
 \\
& Experience replay strategies
& \cite{doro2023sampleefficient},\cite{Mnih2015Humanlevelcontrol},\cite{ICLR2016_PER},\cite{3295258}, \cite{NEURIPS2019_e6d8545d},\cite{3694378} \\
& Prior knowledge-driven 
& \cite{Alharbi_Roozbehani_Dahleh_2024},\cite{pmlr-v177-lu22a},\cite{tirumala2022behavior},\cite{ICLR2024_c1842fcf},\cite{bhambri2024efficient},\cite{yan2025efficient} \\
\hline
\multirow{3}{=}{\centering Resource-Efficient \& Generalizable RL} 
& Few-shot RL
& \cite{3504429}, \cite{pmlr-v139-dance21a}, \cite{daoudi2024conservative} \\
& Meta-RL
& \cite{anantha2020generalized}, \cite{finn2017m}, \cite{duan2016rl2}, \cite{pmlr-v97-rakelly19a} \\
& Computation-limited RL
& \cite{NEURIPS2024_9dcc5038}, \cite{3739782}, \cite{ma2025efficient}, \cite{3693242}, \cite{lee2025hyperspherical}, \cite{pmlr-v291-foster25a} \\
\hline
\multirow{5}{=}{\centering Sparse Reward \& Efficient Exploration in RL} 
& Intrinsic signal–driven exploration 
& \cite{8014804}, \cite{burda2018exploration}, \cite{Alet*2020Meta-learning}, \cite{Raileanu2020RIDE}, \cite{Kayal2025TheIO}, \cite{NIPS2016_afda3322}, \cite{ostrovski2017count}, \cite{pmlr-v202-lobel23a} \\
& Memory and trajectory guided exploration 
& \cite{savinov2018episodic}, \cite{Badia2020Never}, \cite{10347362},
 \cite{ecoffet2019goexplore}, \cite{pmlr-v202-gallouedec23a} \\
& Representation and skill–based exploration
& \cite{eysenbach2018diversity}, \cite{NEURIPS2018_e6384711}, \cite{3327385}, \cite{JMLR:v24:22-0606}, \cite{NEURIPS2020_a322852c}, \cite{pmlr-v139-rybkin21b}, \cite{mai2022sample} \\
\hline
\end{tabular}
\label{tab:reinforcement_literature}
\end{table}

In low-resource scenarios, reinforcement learning research mainly focuses on three core directions: \textbf{Sample-Efficient Reinforcement Learning}, which enables an agent to rapidly learn an optimal or near-optimal decision-making policy under the constraint of extremely limited environmental interaction samples (i.e., experience); \textbf{Low Interaction and Generalization-Efficient Reinforcement Learning}, which aims to balance between cross-task transfer and generalization in low-interaction environments; and \textbf{Sparse Reward Reinforcement Learning}, which seeks to improve learning efficiency through efficient exploration under sparse reward conditions. These directions collectively address the challenges of operating in environments with constrained data, computational resources, or feedback signals. By addressing these constraints, researchers aim to develop robust algorithms capable of effective performance across diverse and complex tasks. Figure \ref{fig:RL-AC} illustrates the typical learning process of reinforcement learning, and in Table \ref{tab:reinforcement_literature}, we provide a classified summary of the latest research findings in the field of low-resource reinforcement learning.

\underline{ \textcircled{1} \textbf{Sample Efficient Reinforcement Learning.}} With the core goal of significantly reducing the requirement for interaction samples, this research topic focuses on studying sample efficiency in reinforcement learning and can be roughly categorized into the following types:
\begin{enumerate}[label=\arabic*)]
  \item \textbf{Model-based methods}. By learning a dynamics model (or world model) of the environment, planning or virtual interaction is conducted within the model to reduce direct interaction with the real environment. Representative methods include Dyna \cite{122377}, Deep Planning Network (PlaNet) \cite{hafner2019learning}, Model-Based Policy Optimization (MBPO) \cite{NEURIPS2019_5faf461e}, Stochastic Ensemble Value Expansion (STEVE)\cite{NEURIPS2018_f02208a0}, Probabilistic Ensemble Trajectory Sampling (PETS)\cite{3327385}, Dreamer\cite{Hafner2020Dream},  DreamerV2\cite{hafner2021mastering}, EfficientZero\cite{NEURIPS2021_d5eca8dc}, H-UCRL\cite{curi2020efficient}.
 \item \textbf{Experience replay strategies}. This type of method enhances sample utilization efficiency and reduces reliance on new samples by improving the experience replay mechanism. Key examples include: Deep Q-Networks, which achieve generalization with high-dimensional inputs through uniform replay \cite{Mnih2015Humanlevelcontrol}; prioritized experience replay \cite{ICLR2016_PER}; hindsight experience replay \cite{3295258}; episodic backward update, which employs full-episode replay \cite{NEURIPS2019_e6d8545d};  reset replay \cite{3694378}; and replay ratio barrier breaking \cite{doro2023sampleefficient}.
 \item \textbf{Prior knowledge-driven}. By incorporating prior knowledge related to system dynamics, \cite{Alharbi_Roozbehani_Dahleh_2024} optimized partial dynamics knowledge; \cite{pmlr-v177-lu22a} introduced causal structure into the standard Markov Decision Processes (MDPs); \cite{tirumala2022behavior} achieved this by probabilistically modeling hierarchical behavior priors, while \cite{ICLR2024_c1842fcf} did so by introducing pre-trained goal priors to regularize high-level policies; \cite{bhambri2024efficient} and \cite{yan2025efficient} extracted domain knowledge, background knowledge, or action distribution priors from LLMs.
\end{enumerate}

\underline{ \textcircled{2} \textbf{Resource-Efficient and Generalizable Reinforcement Learning.}}
These studies are categorized and summarized around three core directions: \textbf{Few-shot RL}, \textbf{Meta-RL} and \textbf{Computation-limited RL}, with details as follows:
\begin{enumerate}[label=\arabic*)]
   \item  \textbf{Few-shot RL}. This category of methods leverages a small number of samples or demonstrations to quickly adapt to new tasks, such as   Deep Q-Learning from Demonstrations (DQfD) \cite{3504429} accelerated learning with small-scale demonstration data, automatically evaluated the proportion of demonstration data; Demonstration-Conditioned RL (DCRL) \cite{pmlr-v139-dance21a} constructed a policy taking demonstrations as input to address issues such as domain shift and suboptimal demonstration improvement in few-shot imitation learning;   Offset Dynamic RL (ODRL)\cite{daoudi2024conservative} used penalty terms to regulate trajectories generated by policies trained in the source environment, enabling policy transfer from the source environment to target environments.
   \item \textbf{Meta-RL}. Meta-RL equips models with rapid cross-task generalization capabilities through multi-task training. Representative works in this area include: the study in \cite{finn2017m} proposed a model-agnostic meta-learning algorithm explicitly supporting adaptation to new tasks using only a small number of gradient updates; $RL^2$ \cite{duan2016rl2} leveraged RNNs to encode fast RL algorithms and trains RNN weights via slow RL, thereby enabling rapid adaptation to new MDPs; the work in \cite{pmlr-v97-rakelly19a} designed an offline meta-reinforcement learning algorithm that decouples task inference from control and integrates offline RL to enhance the efficiency of meta-training and adaptation; and the research in \cite{anantha2020generalized}, which proposed a generalized RL meta-learning framework that explores patterns in the loss surface to optimize few-shot task learners.    
   \item \textbf{Computation-limited RL}. This line of research explores the design of RL algorithms under computational and statistical constraints, aiming to reduce overhead while enhancing efficiency. Recent works address key challenges such as the computational burden of function approximation in distributionally robust offline RL \cite{NEURIPS2024_9dcc5038}, excessive storage and computational costs under Multinomial Logit approximation \cite{3739782}, instability in diffusion policies \cite{ma2025efficient}, high GPU memory usage in RL-based alignment of LLMs \cite{3693242}, and optimization stability in large-scale RL training \cite{lee2025hyperspherical}, while also revealing computational–statistical trade-offs in pre-trained model coverage \cite{pmlr-v291-foster25a}. Collectively, these advances highlight strategies for balancing efficiency, stability, and scalability in computation-constrained RL.
\end{enumerate}

\underline{ \textcircled{3} \textbf{Sparse Reward and Efficient Exploration in Reinforcement Learning.}} To address the issue of inefficient exploration in sparse reward scenarios, researchers have proposed a variety of solutions. Studies related to this topic can be categorized into the following types: 
\begin{enumerate}[label=\arabic*)]
  \item \textbf{Intrinsic signal–driven exploration}. 
  Methods in this category construct intrinsic rewards or frequency-based signals, enabling agents to actively explore even in the absence of extrinsic rewards. Representative works include curiosity- and prediction-driven methods \cite{8014804, burda2018exploration, Alet*2020Meta-learning, Raileanu2020RIDE, Kayal2025TheIO}, as well as counting and pseudo-counting approaches \cite{NIPS2016_afda3322, ostrovski2017count, pmlr-v202-lobel23a}.  
  \item \textbf{Memory and trajectory guided exploration}. 
  These approaches leverage episodic memory or trajectory backtracking to sustain long-term exploration and systematically revisit key states for broader coverage. Representative examples include episodic memory and long-horizon exploration methods \cite{savinov2018episodic, Badia2020Never, 10347362}, as well as staged exploration strategies based on backtracking \cite{ecoffet2019goexplore, pmlr-v202-gallouedec23a}.  
  \item \textbf{Representation and skill–based exploration}.  
  This line of work learns transferable representations or skills through unsupervised learning and latent variable modeling, thereby improving sample efficiency and mitigating distribution shift. Relevant studies include skill discovery and pre-training approaches \cite{eysenbach2018diversity, NEURIPS2018_e6384711}, and latent variable–based exploration \cite{3327385, JMLR:v24:22-0606, NEURIPS2020_a322852c, pmlr-v139-rybkin21b, mai2022sample}.  
\end{enumerate}

\underline{ \textcircled{4}   \textbf{Advances.}}   In the context of RL, leveraging low-resource data holds considerable promise and may play a pivotal role across various strategic paradigms. By effectively utilizing sparse rewards, the models can  reduce the dependence on extensive interactions with the environment. Furthermore, low-resource data can facilitate faster policy adaptation, which supports knowledge transfer to unseen tasks. One promising approach is to leverage prior knowledge from related tasks or domains to compensate for limited data availability in the target task.  For instance, in value-based approaches, low-resource data can be employed to influence the expected return during policy evaluation, thereby facilitating the identification of the optimal policy ${\mathbf{\pi}}^{\ast}$ \cite{zang2020metalight}. Similarly, within policy-based strategies, particularly during direct policy search, low-resource data can serve an auxiliary function, such as perturbing the direction of the policy gradient, which may ultimately influence the trajectory of policy optimization. Additionally, in actor-critic frameworks, low-resource data learning methods can contribute to the refinement of value estimates generated by the critic component \cite{mitchell2021offline}. Collectively, these applications underscore the potential of low-resource data learning as a supportive mechanism to enhance both the efficiency and robustness of RL models through more effective data representation. The integration of low-resource data learning into RL paradigms represents a compelling direction for future research.

\subsection{Hierarchical Structure Modeling with Low-resource Data}

\begin{figure*}
    \centering  
    \subfigure[Modeling of spatial hierarchical structure \cite{106674}.]{\includegraphics[width=0.45\linewidth]{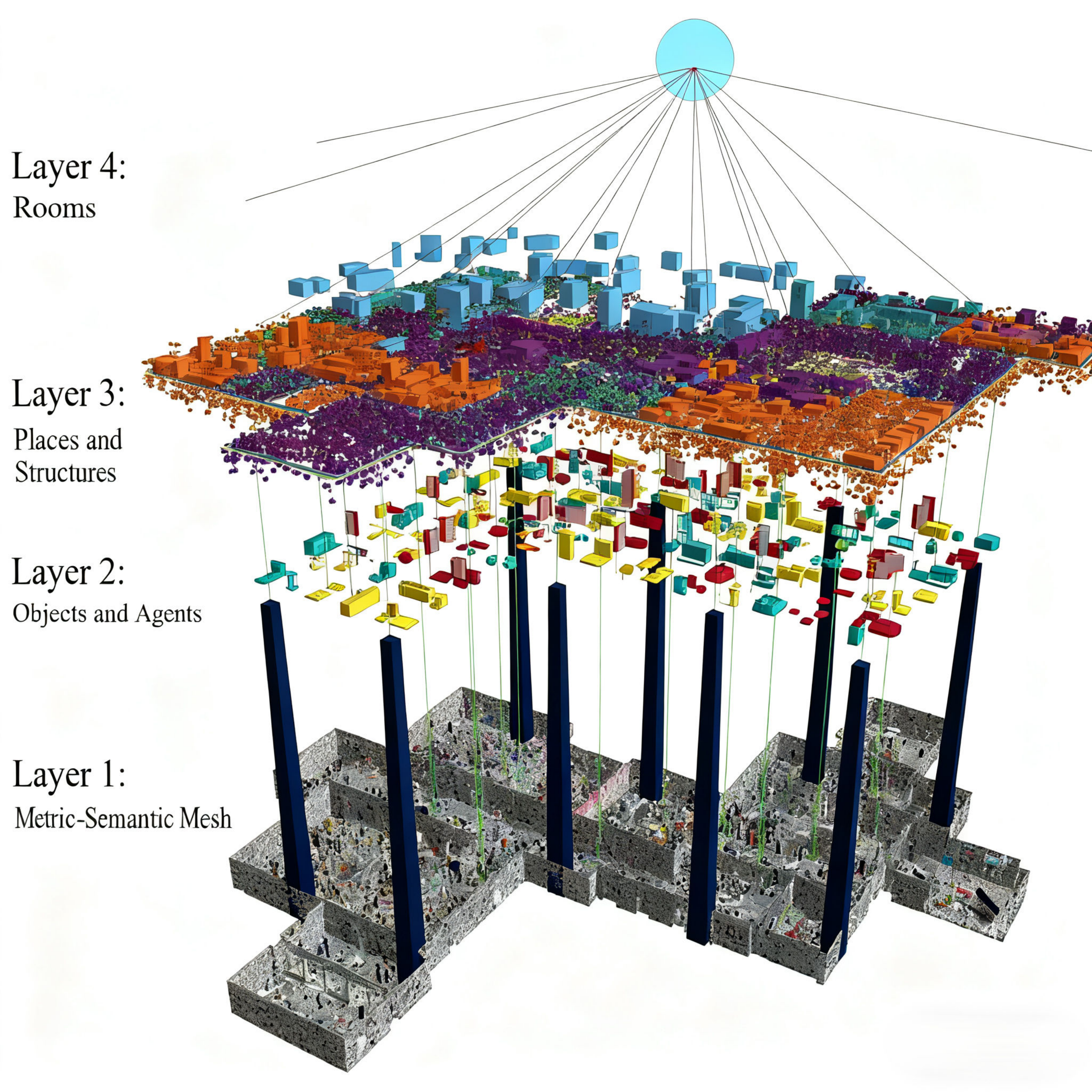} }        
        \label{fig:left} 
    \subfigure[The Poincaré disk model $\mathbb{B}$ is obtained by projecting each point of the hyperboloid model $\mathbb{H}$ onto the hyperplane $\boldsymbol{o}$. ]{\includegraphics[width=0.45\linewidth]{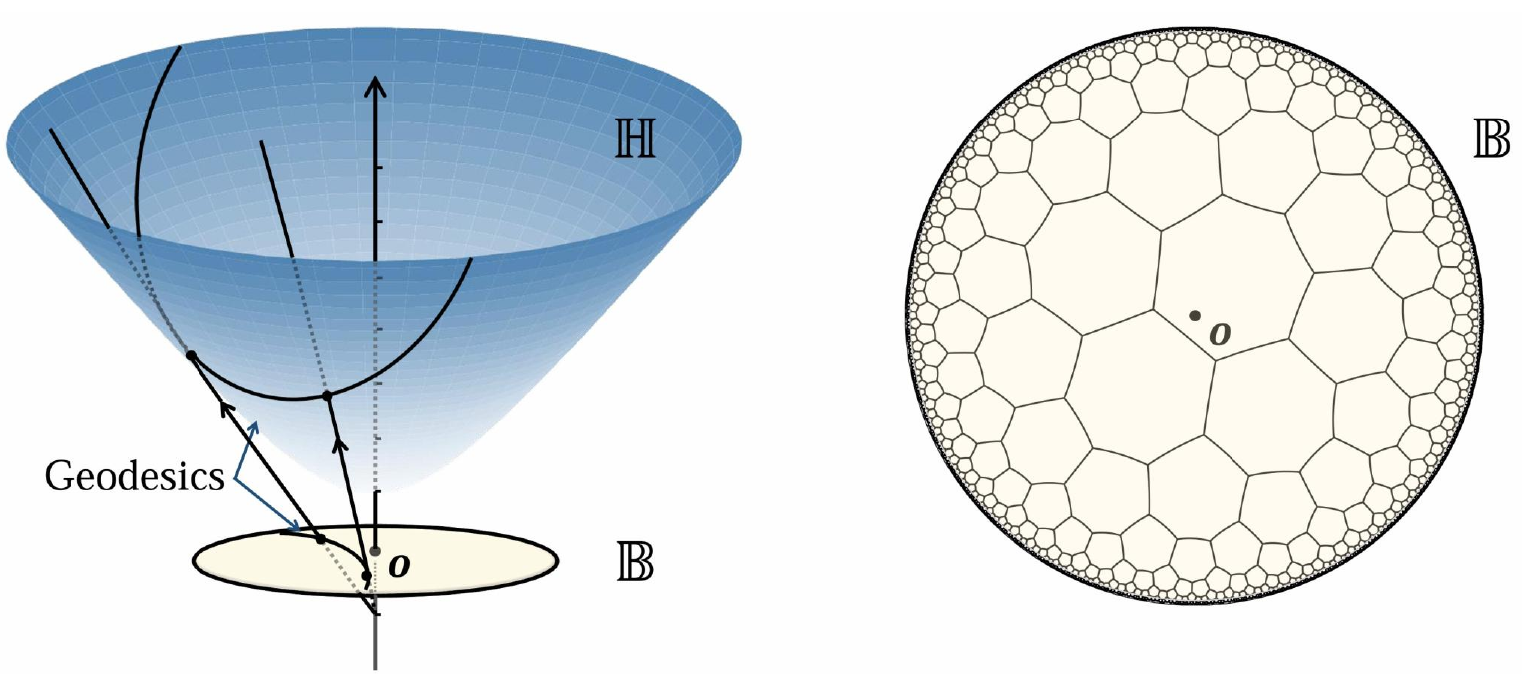}  }
        \label{fig:right}  
    \caption{\cao{Hierarchical Structure Modeling:
    Left figure (a) shows that the spatial structure of embodied intelligent agents executing designated tasks in a home scenario can be modeled as hierarchical-structured data, while right figure (b) presents the hierarchical structure mapping in hyperbolic space from a principled perspective.}} 
    \Description{Hierarchical Structure Modeling}
    \label{fig:HSM}      
\end{figure*}

As a core technology for processing complex topological data, Graph Learning still faces two key challenges: first, the scarcity of labeled data and the high cost of manually annotating graph data; second, the limitation of computing resources. Deep GNNs have a large parameter scale and complex topological computing processes, making them difficult to adapt to low-resource scenarios such as edge devices. In response to these challenges, the following content will sort out the research progress in this field from three aspects: \textbf{Few-Shot Learning on Graphs}, \textbf{ Hierarchical Graph Representation  Learning under Resources Constraints}, and \textbf{Hierarchical Structure Decision Modeling}. In Figure \ref{fig:HSM}, we present examples of hierarchical structures in practical applications and theoretical construction. Meanwhile, in Table \ref{tab:HRM_literature}, we summarize the latest research advances in the field of low-resource hierarchical modeling.

\begin{table}
\centering
\caption{\cao{Classification and Statistics of Research Literature on Hierarchical Structure Modeling under Low-Resource.}}
\begin{tabular}{
  >{\arraybackslash}p{3.5cm} 
  >{\arraybackslash}p{5.0cm} 
  >{\arraybackslash}p{5.0cm}   
} 
\hline
\textbf{Research} & \textbf{Subcategory} & \textbf{Representative Papers} \\ 
\hline
\multirow{5}{=}{\centering Few-Shot Learning on Graphs} 
& Relational reasoning and transfer
& \cite{garcia2018fewshot}, \cite{liu2019fewTPN}, \cite{Kim_2019_CVPR}, \cite{Gidaris_2019_CVPR}, \cite{Chauhan2020FEW-SHOT}, \cite{Yao_Li_2020}, \cite{8954122}, \cite{ding2022meta} \\
& Meta-learning module design
&  \cite{ijcai2022p789}, \cite{NEURIPS2020_c055dcc7}, \cite{Yang_2020_CVPR}, \cite{3411922}, \cite{pmlr-v119-qu20a} \\
& Structural and adaptability enhancement
& \cite{Liu_Fang_Liu_Hoi_2021},   \cite{Chen_2021_CVPR}, \cite{9350663}, \cite{3603091}, \cite{Yu_He_Song_Xiang_2022}, \cite{3539265} \\
& LLM-empowered technologies
& \cite{3715854}, \cite{pmlr-v235-chen24bh},\cite{Hu_Zou2025},\cite{3657775}, \cite{zhou-etal-2025-graph}
\\
\hline
\multirow{3}{=}{\centering Computational Resource Constraints} 
& Lightweight design
& \cite{pmlr-v97-wu19e}, \cite{3294869}, \cite{chen2018fastgcn}, \cite{3330925}, \cite{graphsaint-iclr20}, \cite{yang2023selfsupervised} \\
& Pruning sparsification
& \cite{pmlr-v139-chen21p}, \cite{10164013}, \cite{shin2024feasibility}, \cite{3403236} \\
& Knowledge distillation
& \cite{yang2020distilling}, \cite{ijcai2021p314}, \cite{NEURIPS2022_c06f7889}, \cite{zhang2022graphless}, \cite{3619972} \\
\hline
\multirow{2}{=}{\centering Hierarchical Structure Decision Modeling} 
& Hierarchical RL decision modeling
& \cite{NEURIPS2018_e6384711}, \cite{zhang2021hierarchical}, \cite{pmlr-v202-li23ad}, \cite{JMLR:v26:24-1184}, \cite{pmlr-v235-ma24b}\\ & Large models-related decision modeling
& \cite{fu-etal-2024-msi}, \cite{nottingham2023embodied},    \cite{zhai2024finetuning}, \cite{eigner2024determinants} 
\\
\hline
\end{tabular}
\label{tab:HRM_literature}
\end{table}

\underline{ \textcircled{1}   \textbf{Few-Shot Learning on Graphs.}}
\begin{enumerate}[label=\arabic*)]
\item \textbf{Relational reasoning and transfer.}  
In the early stage, GNNs with message-passing mechanisms \cite{garcia2018fewshot} were used to conduct relational reasoning in semi-supervised and active learning scenarios, thereby alleviating the problem of data scarcity. Transductive Propagation Networks achieved this by learning label propagation on manifold structures \cite{liu2019fewTPN}. Edge-Labeling GNNs \cite{Kim_2019_CVPR} realized explicit clustering through edge prediction, improving the modeling effectiveness both within and between classes; the method proposed in \cite{Gidaris_2019_CVPR} generated classification weights from noisy prototypes via denoising autoencoders integrated with GNNs, and captured the dependencies between categories. The study in \cite{Chauhan2020FEW-SHOT} aggregated base categories into super-classes based on spectral measures, which strengthened the distinguishability of category labels. The approach in \cite{Yao_Li_2020} transferred knowledge from auxiliary graphs by sharing a metric space, enhancing the semi-supervised performance under few annotations. The works in \cite{8954122,ding2022meta} unified semi-supervised learning and propagation methods to improve performance under few-label settings.
\item \textbf{Meta-learning module design.}  
In addition to enabling generalization capabilities across node, edge, or entire graph levels with limited samples \cite{ijcai2022p789}, several frameworks have been developed to enhance task adaptability: MetaTNE facilitated connections between graph structures and new labels via embedding transformation \cite{NEURIPS2020_c055dcc7}; DPGN modeled both instance-level and distribution-level relationships within a dual-graph structure to boost learning performance \cite{Yang_2020_CVPR}; Graph Prototypical Networks extracted meta-knowledge from task pools to support few-shot node classification \cite{3411922}; and relational graph-based Bayesian meta-learning addressed uncertainty in few-shot relation extraction tasks \cite{pmlr-v119-qu20a}.
\item \textbf{Structural and adaptability enhancement. } 
RALE modeled task dependencies using positional embedding \cite{Liu_Fang_Liu_Hoi_2021}; ECKPN fused instance-level and class-level graph features \cite{Chen_2021_CVPR}; and HGNNs extracted discriminative features through multi-level pooling \cite{9350663}. Furthermore, task-specific structures mitigated meta-task differences \cite{3603091}; Hybrid GNNs improved inductive learning performance and reduced sensitivity to outliers \cite{Yu_He_Song_Xiang_2022}; and adaptive modules addressed distribution shift issues under long-tailed distributions \cite{3539265}. 
\item \textbf{LLMs-empowered technologies.} Recent LLMs-empowered research has introduced technologies such as pre-training prompts, which effectively reduced the variability of prompt fine-tuning and enhanced model robustness \cite{3715854}; LLaGA \cite{pmlr-v235-chen24bh} integrated the capabilities of LLMs to handle the complexity of graph-structured data; the work in \cite{Hu_Zou2025} distilled the semantic capabilities of LLMs into more efficient small-scale GNNs, addressing graph learning issues in resource-constrained environments; by introducing graph instruction tuning, the research in \cite{3657775} enabled LLMs to understand graph structures, endowing the models with strong generalization capabilities across various graph learning tasks; the study in \cite{zhou-etal-2025-graph} converted graph structures into language corpora and leveraged the cross-lingual transfer ability of LLMs to enable graph structure understanding.
\end{enumerate}

\underline{ \textcircled{2} \textbf{Hierarchical Graph Representation under Resource Constraints.}}  
Against the backdrop of constrained computing resources, research on hierarchical graph representation has been increasingly centered on addressing efficiency bottlenecks—thereby establishing model compression and efficient inference as core frontiers in this domain.
Recent advances can be structured into three dimensions:   \textbf{Lightweight design}, building compact architectures;  \textbf{Pruning sparsification}, reducing structural redundancy;  and \textbf{Knowledge distillation (KD)}, transferring semantic knowledge across models.  
\begin{enumerate}[label=\arabic*)]
\item \textbf{Lightweight design}. To address the high computational complexity of GNNs, lightweight methods have significantly reduced resource requirements. For instance, the work in \cite{pmlr-v97-wu19e} merged the non-linear operations and multi-layer weights of GCNs into a single linear transformation while preserving predictive performance; \cite{3294869} proposed an inductive low-resource learning framework based on neighbor sampling; \cite{chen2018fastgcn} treated graph convolution as an integral transformation and adopted Monte-Carlo sampling estimation to effectively mitigate the neighbor explosion problem; \cite{3330925} implemented mini-batch training based on graph partitioning for resource-constrained environments; \cite{graphsaint-iclr20} proposed a subgraph sampling training method that significantly accelerates the training process of large-scale graphs; and targeting the limitations of the propagation mechanism, \cite{yang2023selfsupervised} developed LRD-GNN-Matrix based on low-rank decomposition.
\item \textbf{Pruning sparsification}. Pruning techniques reduce the computational burden of GNNs by eliminating redundant structures. The work in \cite{pmlr-v139-chen21p} extended the Lottery Ticket Hypothesis to GNNs, demonstrating that sparse subnetworks can significantly compress the model while preserving performance. \cite{10164013} proposed a progressive pruning and sparse training strategy. Based on fidelity metrics, \cite{shin2024feasibility} explored the feasibility of graph pruning and put forward a new pruning criterion, offering novel insights for efficient GNN design. \cite{3403236} developed an efficient GNN structure that greatly reduces the number of parameters and accelerates both training and inference.
\item \textbf{Knowledge distillation}. Knowledge distillation enables efficient inference by transferring knowledge from complex GNNs to lightweight models. The research in \cite{yang2020distilling} proposed a KD method for GCNs, integrating a topology-preserving module to construct a topology-aware student model. \cite{ijcai2021p314} designed a self-distillation framework for GNNs that eliminates the need for an explicit teacher model. \cite{NEURIPS2022_c06f7889} leveraged a Neural Heat Kernel to capture topological information, enabling KD and compression from complete graphs to sparse graphs. \cite{zhang2022graphless} achieved KD from GNNs to MLPs, incorporating the advantages of MLPs  that have low latency and no graph dependency while preserving the predictive accuracy of GNNs. \cite{3619972} proposed a distillation mechanism based on knowledge reliability evaluation, which selects high-quality nodes as supervision signals to enhance the performance of student MLPs.
\end{enumerate}

\underline{ \textcircled{3}   \textbf{Hierarchical Structure Decision Modeling.}} To address the decision-making requirements of complex tasks, hierarchical structure decompose the process into functional modules such as perception, planning, execution, and feedback, thereby promoting the improvement of the system's intelligence level. Embodied intelligence is a key node for realizing autonomous AI \cite{gupta2021embodied}. As a complex engineering endeavor, its effective implementation is inseparable from the integration of fields such as low-resource \textbf{domain generalization} and \textbf{sample-efficient RL}. At the decision-making level, hierarchical structures provide effective solutions for decision-making and have become a current research focus.
\begin{enumerate}[label=\arabic*)]
    \item \textbf{Hierarchical RL decision modeling.} For instance, the work in \cite{NEURIPS2018_e6384711} introduced Data-Efficient HRL, where higher-level controllers automatically learn and propose goals to supervise lower-level controllers. Based on HRL, HIDIO \cite{zhang2021hierarchical} learned task-agnostic options in a self-supervised manner, thereby improving sample interaction efficiency in sparse-reward scenarios. \cite{pmlr-v235-ma24b} constructed a general sequence modeling framework from the perspective of HRL; within this framework, the high-level policy generates prompts to guide the low-level policy in action generation, and Decision Transformer (DT) emerges as a special case under specific conditions. \cite{pmlr-v202-li23ad} employed a Hierarchical Trajectory-level Diffusion Probabilistic Model for decision-making in offline RL, alleviating the "deadly triad" problem, as well as issues of limited data and reward sparsity. \cite{JMLR:v26:24-1184} leveraged structural information embedded in the decision-making process to automatically construct hierarchical policies in HRL.
    \item \textbf{Large model-related decision modeling.} For example, the work in \cite{fu-etal-2024-msi} introduced the Multi-Scale Insight Agent, which is an embodied agent designed to enhance LLMs' planning and decision-making abilities by effectively summarizing and utilizing insights across different scales. DECKARD \cite{nottingham2023embodied}, an agent for embodied decision-making, used few-shot LLMs to hypothesize an Abstract World Model for RL, significantly improving sample efficiency in Minecraft item crafting. \cite{zhai2024finetuning} proposed fine-tuning vision-language models by leveraging chain-of-thought reasoning and RL environmental feedback to enhance agents' decision-making capabilities. \cite{eigner2024determinants} conducted a comprehensive analysis of the influencing factors in LLMs-assisted decision-making, which holds significant importance for human-AI collaboration.
\end{enumerate}

\underline{ \textcircled{4}   \textbf{Advances.}} Learning with low-resource data not only provides crucial support for graph representation learning, but also enables deep integration with hierarchical structure modeling. By means of hierarchical decomposition, multi-scale representation, and hierarchical optimization of graph structures, it can further enhance the effectiveness of graph models in complex scenarios, while offering underlying graph structure support for hierarchical structure decision modeling. For instance, it can extract effective low-resource data representations in graph contrastive learning to assist in the selection of hard negative samples \cite{xia2022progcl}, and enhance the extraction of meaningful graph representations in various graph mining tasks; in addition, it can filter out untrustworthy or noisy data, helping to build highly generalized and reliable graph representation models \cite{hamilton2017representation}. In terms of graph statistical characteristics, node centrality metrics (including degree centrality, closeness centrality, eigenvector centrality, betweenness centrality, etc.) \cite{8247210,rodrigues2019network,10507035} play a crucial role.

In summary, the collaborative integration of learning with low-resource data, graph representation learning, and hierarchical structure modeling not only enables the exploration of new collaborative application scenarios such as hierarchical graph representation and hierarchical graph mining, but also serves as the underlying technical support for hierarchical structure decision modeling. For example, by characterizing the topological associations in decision-making scenarios (e.g., the graph-structured hierarchy between task objectives and execution steps) through hierarchical graph representation, it can help hierarchical decision-making modules (perception, planning, and execution modules) handle complex tasks more efficiently. The core challenge in combining the two lies in balancing two aspects under the constraint of limited data. First, the accuracy of the graph topology's hierarchical decomposition must be maintained to prevent the loss of key information from overly coarse divisions. Second, the model's computational efficiency is crucial to avoid exacerbating data sparsity and the computational burden caused by an excessive number of hierarchies. This also represents a key direction for future research.

\section{Conclusion}
In this analytical survey, we systematically summarize the analysis and investigation of low-resource learning by synthesizing theoretical insights and scenario-based investigation, and elaborating on the field's future challenges and feasible development trajectory. Specifically, we present a formal theoretical analysis of learning with low-resource data, establishing rigorous guarantees in both supervised and unsupervised settings. We analyze model-agnostic generalization through error bounds and label complexity within the PAC learning framework, showing that near-optimal performance is achievable under data-scarce conditions with controlled loss. Building on these foundations, we examine principled optimization strategies designed for low-resource learning. Ultimately, the analysis concludes by identifying representative scenarios where such strategies improve data representation and generalization, thereby offering a coherent framework that links theoretical guarantees with practical methodologies.

The section \emph{Investigation from Scenario} concludes that organizing vertical explorations of low-resource learning across paradigms, including domain transfer, reinforcement learning and hierarchical decision modeling, provides a unified framework for understanding generalization and efficiency in low-resource scenarios. The benefits lie in clarifying generalization behaviors, improving sample efficiency and exploration strategies, enabling learning under limited data and resources, and supporting scalable, interpretable decision-making, thereby bridging theory and practice in low-resource learning. Overall, the future trajectory of low-resource learning is anticipated to revolve around effective transfer, reliable generalization, and adaptive approaches to parameter tuning, a trend that may contribute to the broader pursuit of autonomous intelligence.

\section*{Acknowledgements}
Part of this work was finished when Dr. Cao was a research assistant at AAII of the University of Technology Sydney. Thanks for the discussion with Weixin Bu and Yaming Guo.




\clearpage
\bibliographystyle{ACM-Reference-Format}
\bibliography{sample-base.bib}


\appendix
\clearpage
\setcounter{section}{1}

\section*{Supplementary Material of The Survey}
\label{SMT}

\subsection*{Part A: Investigation from Optimization}

\subsubsection*{A1 Geometry-Aware Optimization}
\label{A1:GAO}

\par Gradients and their parameter updates usually remain powerful and versatile optimization methods in Euclidean space. Geometry-aware optimization techniques offer specialized advantages in contexts where the data or model structure aligns with non-Euclidean geometries, such as manifold covers, hierarchical structures, graphs, and trees. In these cases, the mean update is a typical approach for optimizing geometric representations.

\paragraph*{A1.1 Fréchet Mean}
To understand a collection of observations sampled from a statistical population, the \textit{mean} of observations is adopted as one powerful statistics to summarize the observations from an underlying distribution. What does \textit{mean} mean? It may vary under different data distributions and depend on the goal of the statistics.  To characterize the representation of real-world data, typical candidates for mean may be the arithmetic mean and the median, but in some cases, the geometric or harmonic mean may be preferable. When the data exists in a set without a vector structure, such as a manifold or a metric space, a different concept of \textit{mean} is required, \ie, Fréchet mean \cite{lee2006riemannian}.

\noindent \textbf{Fréchet mean on probability measures.}  We explore a general mean that can be defined with less structure but encompasses the common notions of \textit{mean} – the \textit{Fréchet mean}. The Fréchet mean is an important entailment (implication) in geometric representations, that embeds a ``centroid”  to indicate its local features (neighbors) on a metric space. For a distance space $(\mathcal{X},d_{\mathcal{X}})$, let $\mathbb{P}$ be a probability measure on $\mathcal{X}$ with $\int d_{\mathcal{X}}^2(x,y)\dif \mathbb{P}(x) < \infty $ for all $y\in\mathcal{X}$, the Fréchet mean is to operate the argmin optimization \cite{lou2020differentiating} of
\begin{equation}
\mu_{\mathcal{X}} = \argmin_{\mu \in \mathcal{X} } \int d_{\mathcal{X}}^2(x,\mu) \dif \mathbb{P}(x).
    \label{FM_pro}
\end{equation}
The Fréchet means defined by the probability measures is more generalized and can be derived to more common objects.\\
\textbf{Advantages of Fréchet mean.} The Fréchet mean w.r.t. Eq.~(\ref{FM_pro}) has two significant advantages  \cite{lou2020differentiating}. 1) It provides a common construction for many well-known notions of \textit{mean} in machine learning and thus implies many interesting properties of data. 2) It provides the notion of \textit{mean} in spaces with less structure than Euclidean spaces, \eg, metric spaces or Riemannian manifolds, thus widening the possibility of adopting machine learning methods in these spaces.\\
\textbf{Fréchet mean on Riemannian manifold.} We next observe the Fréchet mean on manifold.
For an arbitrary Riemannian manifold $\mathcal{M}$ with metric $g_x(\cdot,\cdot)$ that projects the tangent space $\mathcal{T}_x\mathcal{M}\times \mathcal{T}_x\mathcal{M}\rightarrow \mathbb{R}^n$ where $\|v\|_g=\sqrt{g_x(v,v)}$, let $\gamma(t):[a,b]\to \mathcal{M}$ be a geodesics  that stipulates distance  $d_{\mathcal{M}}(\cdot,\cdot)$ as the integral of the first order of the geodesics. For all $x,y\in\mathcal{M}$, the distance $d_{\mathcal{M}}(x,y):={\rm inf} \int_a^b \|\gamma '(t)\|_g\dif t$ where $\gamma(t)$ denotes any geodesics such that $\gamma(a)=x,\gamma(b)=y$. Given a point set $\mathcal{X}=\{x_1,x_2,...,x_m\}\subseteq\mathcal{M}$ and set the probability density of each point $x_i\in\mathcal{X}$ as $\frac{1}{m}$, the \textit{weighted Fréchet mean} \cite{lou2020differentiating} is to operate the argmin optimization  of
\begin{equation}
\begin{split}
\mu_\mathcal{M} = \argmin_{\mu \in \mathcal{M} } \sum_{i=1}^{m} \omega_i   d^2_\mathcal{M}(x_i,\mu),
\end{split}
\label{argmin}
\end{equation}
where  $\omega_i$ denotes the weight of $x_i$, and the constraint of $\mu \in \mathcal{M}$ stipulates that $\mu$ may converge in $\mathcal{M}$ with infinite candidates. Given  $d_\mathcal{M}(a,b):=\|a-b\|_2$ defined in Euclidean geometry and $\omega_i=1/m$ for all $i$, the weighted Fréchet mean is then simplified into the Euclidean mean, which results in a fast computational time. This specified setting  achieved promising results in the $k$-means clustering, maximum mean discrepancy optimization \cite{gretton2012kernel}, \emph{etc.}

\begin{figure*} 
  \centering
  \subfigure[Poincaré disk model and geodesics]{\includegraphics[width=0.238\textwidth]{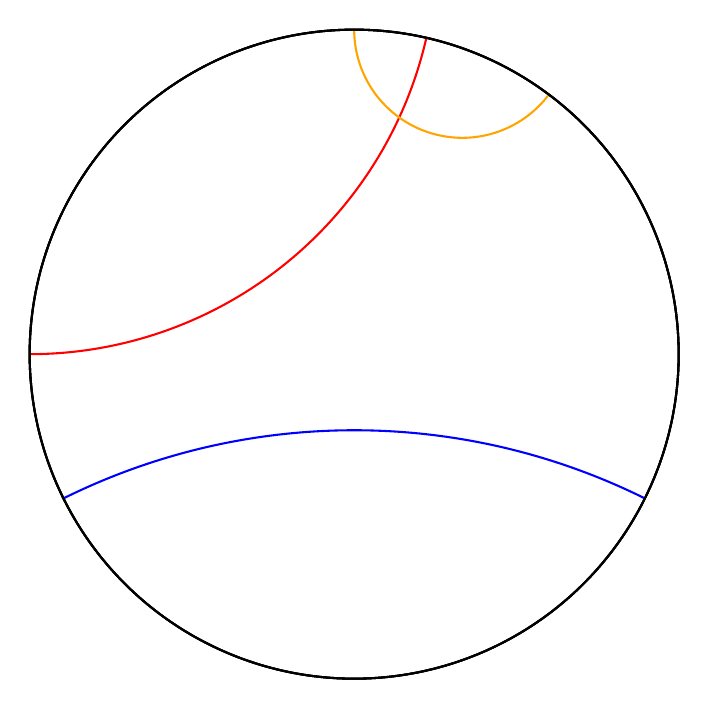}
    \label{fig:poincare}
  }\hspace{6mm}
  \subfigure[Poincaré ball model]{
    \includegraphics[width=0.25\textwidth]{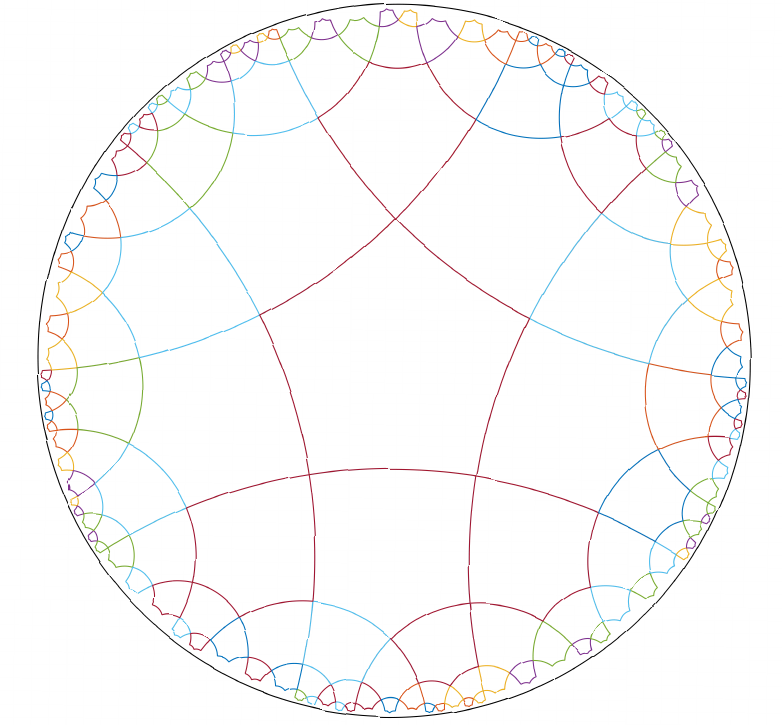}
      \label{fig:poincare1}
  }\hspace{3mm}
  \subfigure[Lorentz model]{
    \includegraphics[width=0.34\textwidth]{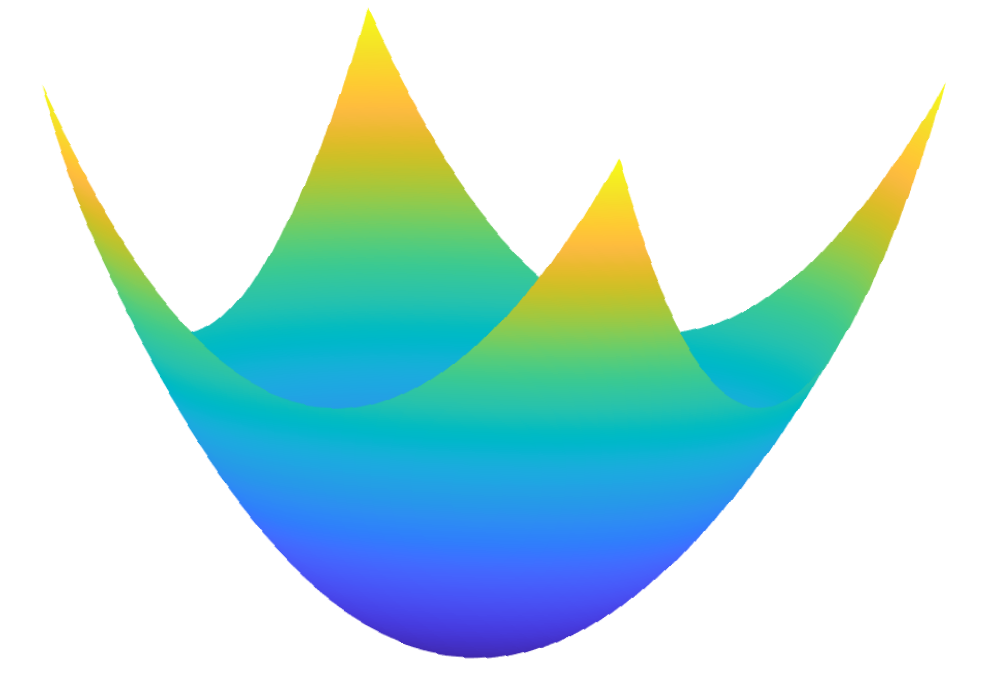}
      \label{fig:lorentz}
  }
  \caption{Illustration of Poincaré (disk and ball) model and Lorentz model. (a) Poincaré disk model (2D Poincaré ball model) and geodesics. Poincaré disk model is an open unit circle and all of its geodesics are perpendicular to the boundary. (b) Poincaré ball model is an open unit ball. The closer two points inside the model are to the boundary, the more the distance between them grows infinitely. (c) Lorentz model. It is defined as the upper sheet of a two-sheeted n-dimensional hyperboloid.}
  \Description{Illustration of Poincaré (disk and ball) model and Lorentz model.}
  \label{fig:two-models}
\end{figure*}

\paragraph*{A1.2 Euclidean Mean}
The Euclidean mean is the most widely adopted \textit{mean} in machine learning. Meanwhile, the Euclidean mean is of great importance to perform aggregation operations such as attention \cite{vaswani2017attention}, 
batch normalization \cite{bjorck2018understanding}. 
Let $\mathcal{R}^n$ denote the Euclidean manifold with zero curvature, and the corresponding Eculidean metric of which is defined as $g^{E}=\mathrm{diag}([1,1, \ldots, 1])$. For $\textbf{x}, \textbf{y} \in\mathcal{R}^n$, the Euclidean distance is given as:
\begin{equation}
    d_{\mathcal{R}}(\textbf{x}, \textbf{y})=\| \textbf{x}-\textbf{y} \|_2.
\end{equation}
Then $(\mathcal{R}^n,d_{\mathcal{R}})$ is a complete distance space. 
We next present a formal description of Euclidean mean \cite{aloise2009np} based on the weighted Fréchet mean w.r.t. Eq.~(\ref{argmin}).

\begin{proposition} \label{proposition-em}
Given a set of points $\mathcal{X}=\{x_1,x_2,....,x_m\}\subseteq\mathcal{R}^n$, the Euclidean mean $\mu_{\mathcal{R}}$ minimizes the following problem:
\begin{equation}\label{equ:euclidean mean}
\begin{split}
\min_{\mu\in \mathcal{R}^{n} } \sum_{i=1}^{m} \omega_i         d^2_\mathcal{R}(x_i,\mu),
\end{split}
\end{equation}
where $\omega_i\geq0$ denotes the weight coefficient of $x_i$. 
\end{proposition}
The completeness of $(\mathcal{R}^{n},d_{\mathcal{R}})$ guarantees that Eq.~(\ref{equ:euclidean mean}) has a closed gradient, so a unique solution exists. With Proposition~\ref{proposition-em}, the Euclidean mean $\mu_{\mathcal{R}}$ is unique with the closed-form solution of
\begin{equation}
    \mu_{\mathcal{R}}=\frac{1}{m}\sum_{i=1}^m x_i.
\end{equation}

\paragraph*{A1.3 Non-Euclidean Mean}
Recent studies demonstrate that hyperbolic geometry has stronger expressive capacity than the Euclidean geometry to model hierarchical features \cite{chami2019hyperbolic,liu2019hyperbolic}. Meanwhile, the Euclidean mean extends naturally to the Fréchet mean on hyperbolic geometry. Then, we discuss the Fréchet mean on the Poincaré  and  Lorentzian model with respect to Riemannian manifold. The illustrations of Poincaré  and Lorentz model are presented in Figure~\ref{fig:two-models}.

\emph{Poincaré Centroid}  \noindent The Poincaré ball model $\mathcal{P}^{n}$ with constant negative curvature corresponds to the Riemannian manifold $(\mathcal{P}^n$, $\left.g_x^{\mathcal{P}}\right)$ \cite{peng2021hyperbolic}, where $\mathcal{P}^n=\left\{x \in \mathbb{R}^{n}:\|x\|<1\right\}$ denotes an open unit ball defined as the set of $n$-dimensional vectors whose Euclidean norm are smaller than $1$. The Poincaré metric  is defined as $g_{x}^{\mathcal{P}}=\lambda_{x}^{2} g^{E}$, where $\lambda_{x}=\frac{2}{1-\|x\|^{2}}$ denotes the conformal factor and $g^{E}$ denotes the Euclidean metric. For any $\textbf{x},\textbf{y}$ in $\mathcal{P}^n$, the Poincaré distance is given as \cite{nickel2017poincare}:
\begin{equation}
   d_{\mathcal{P}}(\textbf{x}, \textbf{y})=\cosh ^{-1}\left(1+2 \frac{\|\textbf{x}-\textbf{y}\|^{2}}{\left(1-\|\textbf{x}\|^{2}\right)\left(1-\|\textbf{y}\|^{2}\right)}\right). 
\end{equation}
Then $(\mathcal{P}^n,d_{\mathcal{P}})$ is a distance space. We thus have the following proposition for the Poincaré centroid \cite{nickel2017poincare} based on the Poincaré distance.

\begin{proposition}
Given a set of points $\mathcal{X}=\{x_1,x_2,....,x_m\}\subseteq\mathcal{P}^n$, the Poincaré centroid $\mu_{\mathcal{P}}$ minimizes the following problem:
\begin{equation}
    \min_{\mu\in \mathcal{P}^{n}} \sum_{i=1}^m \omega_i d^2_\mathcal{P}(x_i,\mu),
\end{equation}
where $\omega_i\geq0$ denotes the weight coefficient of $x_i$.
\end{proposition}
There is no closed-form solution for the Poincaré centroid $\mu_\mathcal{P}$, so Nickel et al. \cite{nickel2017poincare} compute it via gradient descent.

 \emph{Lorentzian Centroid.} \noindent   The Lorentz model $\mathcal{L}^{n}$ \cite{peng2021hyperbolic} with constant curvature $-1 / K$ avoids numerical instabilities that arise from the fraction in the Poincaré metric,  
for $\textbf{x}, \textbf{y} \in \mathbb{R}^{n+1}$, the Lorentzian scalar product is formulated as \cite{law2019lorentzian}:
\begin{equation}
\left<\textbf{x}, \textbf{y} \right>_\mathcal{L} = -x_0y_0 + \sum_{i=1}^n x_iy_i \leq -K.
\end{equation}
This model of $n$-dimensional hyperbolic space corresponds to the Riemannian manifold ($\mathcal{L}$, $g^\mathcal{L}_\textbf{x}$), 
where $\mathcal{L} = \{\textbf{x} \in \mathbb{R}^{n+1} : \left<\textbf{x}, \textbf{y} \right>_\mathcal{L} = -K, x_0 > 0\}$ (\textit{i.e.}, the upper sheet of a two-sheeted n-dimensional hyperboloid)
and $g^\mathcal{L}_\textbf{x} = \mathrm{diag}([-1, 1, ..., 1])$ denotes the Lorentz metric. The squared Lorentz distance for $\textbf{x}, \textbf{y} \in \mathbb{R}^{n+1}$ which satisfies all the axioms of a distance other than the triangle inequality is defined as \cite{law2019lorentzian}:
\begin{equation}
\begin{split}
d_\mathcal{L}^2(\textbf{x}, \textbf{y}) = \|\textbf{x}-\textbf{y}\|^2_\mathcal{L} = -2K - 2\left<\textbf{x}, \textbf{y}\right>_\mathcal{L}.
\end{split}
\end{equation}
Proposition~\ref{Lorentzian-Centroid} presents Lorentz centroid that represents the aspherical distributions under a Lorentz model\cite{law2019lorentzian}. 

\begin{proposition}\label{Lorentzian-Centroid} Given a set of points $\mathcal{X}=\{x_1,x_2,....,x_m\}\subseteq\mathcal{L}^n$, the Lorentzian centroid $\mu_{\mathcal{L}}$ minimizes the following problem:
\begin{equation}
    \min_{\mu\in \mathcal{L}^{n}} \sum_{i=1}^m \omega_i d^2_\mathcal{L}(x_i,\mu),
\end{equation}
where $\omega_i\geq0$ denotes the weight coefficient of $x_i$.
\end{proposition}
The Lorentzian centroid $\mu_{\mathcal{L}}$ is unique with the closed-form solution of
\begin{equation}\label{eq:lorentzian-center}
\mu_{\mathcal{L}} = \sqrt{K} \frac{\sum_{i=1}^m \omega_i x_i}{\left|\|\sum_{i=1}^m \omega_i x_i\|_\mathcal{L}\right|},
\end{equation}
where $\left|\|\mathbf{a}\|_{\mathcal{L}}\right|=\sqrt{\left|\|\mathbf{a}\|_{\mathcal{L}}^{2}\right|}$ denotes the modulus of the imaginary Lorentzian norm of the positive time-like vector $\mathbf{a}$.
\emph{Lorentzian Focal Point} \noindent With \cite{9468944}, the Euclidean norm of Lorentz centroid $\mu_{\mathcal{L}}$ decreases, thus yielding an effective approximation to the focal point which is more representative than Lorentz centroid for the aspherical distributions. However, the approximation cannot only depend on $K$ due to uncertain parameter perturbations. We should also control the coefficient $\omega_i$ to approximate the Lorentzian focal point. Here $\omega_i$ w.r.t. Eq.~(\ref{eq:lorentzian-center}) can be written as \cite{9468944}:
\begin{equation}\label{eq:vi}
\begin{split}
\omega_i = \frac{d^2_\mathcal{L}(x_i, \mu)}{\sum_{i=1}^m d^2_\mathcal{L}(x_i, \mu)},
\end{split}
\end{equation}
Then we present the approximation of Lorentzian focal point in Proposition \ref{Lorentzian-Focal-Point} \cite{9468944}.
\begin{proposition}\label{Lorentzian-Focal-Point}
Given a set of points $\mathcal{X}=\{x_1,x_2,....,x_m\}\subseteq\mathcal{L}^n$, the Lorentzian focal point $\mu_{\mathcal{F}}$ minimizes the following problem:
\begin{equation}
   \min _{\mu \in \mathcal{L}^{n}} \sum_{i=1}^m \omega_i \left<x_i, \mu \right>_\mathcal{L}.
\end{equation}
Then the Lorentzian focal point $\mu_{\mathcal{F}}$ can be approximately given as:
\begin{equation}
\mu_{\mathcal{F}} = \sqrt{K} \frac{\sum_{i=1}^m \omega_i x_i}{|\|\sum_{i=1}^m \omega_i x_i\|_\mathcal{L}|},
\end{equation}
where $\omega_i \geq 0$ follows Eq.~(\ref{eq:vi}). 
\end{proposition}

Modeling preference.  (1) For geometry optimization, structured mean representation are employed due to their effectiveness on manifold and graph data, such as biological networks and protein structures. Within this optimization paradigm, the Poincaré and Lorentz models are two typical non-Euclidean prototypes that enable geometrically consistent optimization over Riemannian gradient.
(2)  From a geometric perspective, the Poincaré ball offers better interpretability and visualization, which supports model understanding, particularly in low-resource training scenarios. Additionally, it is computationally lighter and tends to exhibit fewer numerical stability issues.
(3)  In contrast, the Lorentz model typically excels in large-scale settings with high-dimensional features and is well-suited for capturing deeply hierarchical structures. When appropriate geometric mean representations are used in high-resource data environments, the Lorentz model can also operate more efficiently.

\paragraph*{A1.4 Kernel Mean} 
The kernel mean \cite{muandet2016kernel} could be generalized in Euclidean and hyperbolic geometry, which presents a kernel expression for the geometric mean with respect to  the probability measures of Fréchet mean. 

We first review some properties of reproducing kernel Hilbert space (RKHS). Let $\mathcal{H}$ denote a RKHS over $\mathcal{X}$, then every bounded linear functional is given by the inner product with a unique vector in $\mathcal{H}$ \cite{alt2016linear}. For any $x\in\mathcal{X}$, there exists a unique vector $k_x\in\mathcal{H}$ such that $f(x)=\Braket{f, k_x}$ for every $f \in \mathcal{H}$. The function $k_x=K(x,\cdot)$ is called the \textit{reproducing kernel} for point $x$, where $K(x_1,x_2):\mathcal{X}\times\mathcal{X}\to\mathbb{R}$ is positive definite. For any $k_x,k_y$ in $\mathcal{H}$, the Hilbert distance is given as \cite{gretton2012kernel}:
\begin{equation}
    d_{\mathcal{H}}(k_x,k_y)=\|  k_x-k_y \| = \sqrt{\Braket{k_x-k_y,k_x-k_y}}.
\end{equation}
Then $(\mathcal{H},d_{\mathcal{H}})$ is a complete distance space. Following Eq.~(\ref{FM_pro}), Proposition~\ref{pro:kernel mean} gives a formal description of kernel mean \cite{muandet2016kernel}.
\begin{proposition}\label{pro:kernel mean}
Given a separable RKHS $\mathcal{H}$ endowed with a measurable reproducing kernel $K(x_1,x_2):\mathcal{X}\times\mathcal{X}\to\mathbb{R}$ such that $\int d_{\mathcal{H}}^2(k_x,k_y)\dif \mathbb{P}(x) < \infty$ for all $y\in\mathcal{X}$, where $\mathbb{P}$ denotes a probability measure on $\mathcal{X}$. Then the kernel mean $\mu_{\mathcal{H}}$ minimizes the following problem:
\begin{equation}\label{equ:Kernel Fréchet mean}
        \min_{\mu \in \mathcal{H}} \int d_{\mathcal{H}}^2(k_x,\mu) \dif \mathbb{P}(x).
\end{equation}
\end{proposition}
Based on the completeness of the distance space $(\mathcal{H},d_{\mathcal{H}})$, the following theorem gives the solution of kernel mean, which is consistent with the classical kernel mean defined in \cite{muandet2016kernel}.
\begin{theorem}\label{thm:kernel mean}
The kernel mean $\mu_{\mathcal{H}}$ is unique with the closed-form solution:
\begin{equation}
    \mu_{\mathcal{H}}=\int K(x,\cdot) \dif \mathbb{P}(x),
\end{equation}
where $K(x,\cdot)$ indicates that the kernel has one argument fixed at $x$ and the second free.
\end{theorem}
Theorem~\ref{thm:kernel mean} includes kernel mean in Fréchet mean for the first time, thus maintaining formal uniformity with other standard means, \textit{e.g.} Euclidean mean. The detailed proof of Theorem~\ref{thm:kernel mean} can be found in Part B of the Supplementary Material.

\subsubsection*{A2 LLMs-Powered Optimization}

Optimization problems are core challenges across numerous fields, while traditional methods often rely on extensive data, computational resources, or human intervention. In recent years, LLMs with their natural language understanding and generation capabilities, have provided a new perspective for optimization tasks, particularly demonstrating unique advantages in low-resource, low-human-involvement, and few-shot scenarios. This section reviews the latest advancements of LLMs in the optimization domain, categorized into four aspects: gradient optimization, prompt optimization, evolutionary strategy generation optimization, and scientific problem optimization.
   
\paragraph*{A2.1 LLMs Optimize Gradient}

Inspired by numerical gradient descent, \citet{pryzant2023automatic} used small batch data to generate natural language gradients that critique current prompts and propagate edits in the opposite semantic direction, achieving up to a 31\% performance improvement in initial prompts. \citet{guo2024llm} proposed using LLMs as high-level guidance based on the previous phase of gradient optimization, generating potential improved solutions as restart points for the next phase of optimization. The authors validates the effectiveness of this optimization framework on prompt tuning tasks. \citet{tang2025unleashing} introduced GPO, a gradient-like prompt optimizer that refines task prompts based on update directions and methods derived from parameter learning principles.

\paragraph*{A2.2 LLMs Optimize Prompt}

LLMs have significantly advanced prompt optimization, leading to more effective task performance across various domains.  For example, \citet{yang2024large} introduced a Optimization by PROmpting (OPRO) method that enables LLMs to use meta-prompts for iteratively generating and refining solutions. This approach excels in tasks such as linear regression, the traveling salesman problem, and few-shot prompt optimization, consistently outperforming human-designed prompts. Building on iterative optimization, \citet{nie2023importance} proposed a sequential prompt optimizer that synthesizes directed feedback from historical optimization trajectories, achieving greater stability and efficiency in tasks ranging from maximizing mathematical functions to optimizing poetry prompts. For vision-language models, \citet{liu2024language} developed a black-box natural language prompt optimization method. By evaluating prompt performance and leveraging LLM-driven textual feedback, this method surpasses both human-designed and LLM-generated prompts in challenging one-shot image classification scenarios. \citet{guo2024connecting} combined LLMs with evolutionary algorithms in EvoPrompt, a framework for discrete prompt optimization. Starting from an initial prompt population, EvoPrompt uses evolutionary operators to iteratively generate new prompts, automating the optimization process.  \citet{song2024position} emphasized the role of free-form text in enhancing task understanding, leveraging Transformer-based models to design superior optimization strategies and improve performance predictions in unseen search spaces. 

LLMs have driven significant progress in prompt optimization by enabling iterative, feedback-driven, and evolution-inspired strategies that consistently outperform human-designed prompts. These methods span diverse applications—from mathematical optimization and natural language processing tasks to vision-language modeling, highlighting LLMs' ability to enhance task understanding and adapt to complex, unseen scenarios.

\paragraph*{A2.3 LLMs Optimize Evolution}  

LLMs have significantly advanced evolutionary strategies by enabling automated, efficient, and innovative optimization approaches. For instance, \citet{liu2025large} leveraged LLMs to design multi-objective evolutionary algorithms (MOEAs), using prompt engineering to directly employ LLMs as black-box search operators in a zero-shot manner for decomposition-based MOEAs. Similarly, \citet{pluhacek2023leveraging} utilized LLMs like GPT-4 to generate novel hybrid swarm intelligence optimization algorithms, exploring the automation of metaheuristic design. Further integrating LLMs with evolutionary computation, \citet{liu2024evolution} introduced evolution of heuristics, a novel paradigm combining LLMs and evolutionary computing to enable automated heuristic design. Complementing this, \citet{ye2024reevo} proposed  Language Hyper-Heuristics (LHHs), which harness LLMs to generate heuristic strategies, and introduced the reflection evolution method to enhance LHHs’ capabilities in solving combinatorial optimization problems. Additionally, \citet{lange2024large} explored LLMs’ potential to implement evolutionary optimization algorithms in principle, developing a novel prompting strategy that enables users to derive an LLM-based evolutionary strategy. \citet{brahmachary2025large} introduced the Language-based Evolutionary Optimizer (LEO), a population-based LLM approach that demonstrates zero-shot optimization capabilities across diverse scenarios, including multi-objective and high-dimensional problems.  The studies demonstrated above show how LLMs can drive the automation and enhancement of evolutionary strategies, from multi-objective optimization to heuristic and metaheuristic design, revolutionizing the field of evolutionary computation.

\paragraph*{A2.4 LLMs Optimize Science}

LLMs are becoming increasingly important in scientific optimization, particularly in mathematical, combinatorial, and numerical domains. For example, \citet{guo2023towards} evaluated the optimization capabilities of LLMs across various mathematical and combinatorial tasks, demonstrating their effectiveness as black-box optimizers. LLMs excel in small-scale problems with limited data, though their performance is constrained by problem dimensionality and numerical complexity. \citet{huang2024exploring} conducted a comprehensive study on LLMs' potential in both discrete and continuous optimization. While LLMs perform poorly in purely numerical tasks, they show significant improvement in non-numerical domains when enhanced by heuristic prompts. Furthermore, \citet{pmlr-v235-ahmaditeshnizi24a} introduced the OptiMUS framework, which constructed and solved mixed-integer linear programming problems from natural language descriptions, streamlining complex problem-solving. The study behind LLM4ED \citep{du2024llm4edlargelanguagemodels} proposed a new framework that used LLM-based prompts to automatically extract control equations from data, employing LLMs with two alternating iterative strategies to collaboratively optimize the generated equations.  \citet{gao2024strategyllm} proposed StrategyLLM, enabling LLMs to perform inductive and deductive reasoning, resulting in few-shot prompts that consistently outperform human annotations in mathematical, commonsense, algorithmic, and symbolic reasoning tasks.  


In summary, LLMs are redefining the landscape of optimization by introducing novel strategies grounded in language-based reasoning, feedback synthesis, and evolutionary paradigms. Across gradient-inspired approaches, LLMs can simulate semantic update directions to refine prompts and guide optimization phases. In prompt optimization, LLMs employ iterative, feedback-driven, and meta-evolutionary techniques that consistently outperform human-crafted solutions across diverse tasks, including natural language processing, vision-language modeling, and symbolic reasoning. In the realm of evolutionary strategies, LLMs act as automated generators of heuristic and metaheuristic algorithms, facilitating zero-shot and multi-objective optimization. Furthermore, in scientific domains, especially under low-resource or low-supervision constraints, LLMs demonstrate strong potential as black-box optimizers. They can effectively handle mathematical, combinatorial, and even some numerical tasks through natural language interfaces. Collectively, these advancements position LLMs as versatile and adaptive optimizers, capable of bridging data-driven computation with high-level semantic reasoning in complex optimization scenarios.

\subsection*{Part B: Proofs}

\subsubsection*{B1 Proof of Theorem~\ref{Unsupervised-Fashion}}
\begin{proof} 
IWAL denotes a set of observations for its weighted sampling: $\mathcal{F}_t=\{(x_1,y_1,Q_1), (x_2,y_2,Q_2),..., (x_t,y_t,Q_t)\}$.  {The key step to prove Theorem~\ref{Supervised-Fashion} is to observe the  martingale difference sequence for any pair $f$ and $g$ in the t-time hypothesis class $\mathcal{H}_t$, namely, $\xi_t=\frac{Q_t}{p_t}\Big(\ell(f(x_t),y_t)-\ell(g(x_t),y_t)\Big)-\Big(R(h)-R(g)\Big)$, where $f, g \in \mathcal{H}_t$.}  {By adopting Lemma~3 of \cite{kakade2008generalization}, with $\tau>3$ and $\delta>0$, we firstly know
\begin{equation}
\begin{split}
&{\rm var}[\xi_t|\mathcal{F}_{t-1}]\\
&\leq \mathbb{E}_{x_t}\Big[ \frac{Q_t^2}{p_t^2}\Big(\ell(f(x_t),y_t)   -\ell(g(x_t),y_t)\Big)\!-\!\Big(R(h)-R(g)\Big)^2 |\mathcal{F}_{t-1}\Big]\\
& \leq \mathbb{E}_{x_t}\Big[ \frac{Q_t^2p_t^2}{p_t^2}  |\mathcal{F}_{t-1}\Big]\\
& = \mathbb{E}_{x_t}\Big[ p_t |\mathcal{F}_{t-1}\Big],\\
\end{split}
\end{equation}
and then there exists
\begin{equation}
\begin{split}
&|\sum_{t=1}^T\xi_t| \\
&\leq   \max_{\mathcal{H}_i, i=1,2,...,k} \Bigg\{ 2 \sqrt{\sum_{t=1}^{\tau}\mathbb{E}_{x_t}[p_t|\mathcal{F}]_{t-1}}, 6\sqrt{{\log}\Big(\frac{8{\rm log}\tau }{\delta}\Big)} \Bigg\}  
 \times \sqrt{{\rm log}\Big(\frac{8{\rm log}\tau}{\delta}\Big)},\\
\end{split}
\label{xi_obsevation}
\end{equation}
where $\mathbb{E}_{x_t}$ denotes the expectation over the operation on $x_t$. With Proposition~2 of \cite{cesa2008improved}, we have
\begin{equation}
\begin{split}
&\sum_{t=1}^{\tau}\mathbb{E}_{x_t}[p_t|\mathcal{F}_{t-1}]\\
&\leq \Big(\sum_{t=1}^\tau p_t \Big)+36 {\rm log}\Big(\frac{(3+\sum_{t=1}^\tau)\tau^2}{\delta}\Big) +2\sqrt{\Big(\sum_{t=1}^\tau p_t\Big){\rm log}\Big(\frac{(3+\sum_{t=1}^\tau)\tau^2}{\delta}\Big)}\\
& \leq  \Bigg( \sum_{t=1}^\tau p_t + 6 {\rm log}\Big(\frac{(3+\sum_{t=1}^\tau)\tau^2}{\delta}\Big) \Bigg).   \\
\end{split}
\label{F_obsevation}
\end{equation}}
Then, introducing Eq.~(\ref{F_obsevation}) to  Eq.~(\ref{xi_obsevation}), with a probability at least $1-\delta$, we have

\begin{equation}
\begin{split}
& |\sum_{t=1}^T\xi_t|\\
& \leq  \max_{\mathcal{H}_i, i=1,2,...,k} \Bigg\{\frac{2}{\tau}  \Bigg[\sqrt{\sum_{t=1}^{\tau}p_t}+6\sqrt{{\rm log}\Big[\frac{2(3+\tau)\tau^2}{\delta}\Big] } \Bigg] \times \sqrt{{\rm log}\Big[\frac{16\tau^2|\mathcal{H}_i|^2 {\rm log}\tau}{\delta}\Big]}\Bigg\}.
\end{split}
\end{equation}
{For any $\mathcal{B}_i$, the final hypothesis converges into $h_{\tau}$, which usually holds a tighter disagreement to the optimal $h^*$. We thus have $R(h_\tau)-R(h^*) \leq |\sum_{t=1}^\tau\xi_t|$. Therefore, the error disagreement bound of Theorem~\ref{Supervised-Fashion} holds.} We next prove the label complexity bound of MHEAL. Following Theorem~\ref{Supervised-Fashion}, there exists
\begin{equation}
\begin{split}
& \mathbb{E}_{x_t}[p_t|\mathcal{F}_{t-1}]\\
& \leq 4\theta_{\rm IWAL}K_\ell  \times \Bigg(R^*+\sqrt{(\frac{2}{t-1}){\rm log}(2t(t-1)|)  \frac{|\mathcal{H}|^2  }{\delta})}\Bigg), \\
\end{split}
\end{equation}
where  $R^*$ denotes $R(h^*)$. Let  
$\sqrt{(\frac{2}{t-1}){\rm log}(2t(t-1)|)  \frac{|\mathcal{H}|^2  }{\delta})} \propto O\Big({\rm log}(\frac{\tau|\mathcal{H}|}{\delta})\Big)$, with the proof of Lemma~4 of \cite{cortes2020region}, Eq.~(\ref{F_obsevation}) can thus be approximated  as
\begin{equation}
\begin{split}
& \mathbb{E}_{x_t}[p_t|\mathcal{F}_{t-1}] \\
& \leq 4\theta_{\rm IWAL}  K_\ell\Bigg(\tau R^*\!+\!O\Big(R(h^*)\tau {\rm log}(\frac{t|\mathcal{H}|}{\delta})   \Big )  \!+\! O\Big({\rm log}^3(\frac{t|\mathcal{H}|}{\delta})\Big) \Bigg).
\end{split}
\end{equation}
For any cluster $\mathcal{B}_i$, by adopting the proof of Theorem~2 of \cite{cortes2020region}, we know
\begin{equation}
\begin{split}
& \mathbb{E}_{x_t}[p_t|\mathcal{F}_{t-1}]\\  
&\leq   8 K_\ell\Bigg\{\Big[\sum_{j=1}^{N_i} \theta_{\rm MHEAL} R_j^*\tau p_j\Big] 
 +\sum_{j=1}^{N_i} O\Bigg(\sqrt{R_j^*\tau p_j{\rm log}\Big[\frac{\tau|\mathcal{H}_i|N_i}{\delta} \Big]}\Bigg)\! +\!O\Bigg(N_i {\rm log}^3\Big(\frac{\tau|\mathcal{H}_i|N_i}{\delta}\Big)\Bigg)\Bigg\}. \\
\end{split}
\end{equation}
Since $\mathcal{Q}=k\times \max_{\mathcal{H}_i}\mathbb{E}_{x_t}[p_t|\mathcal{F}_{t-1}], {\rm s.t.}, i=1,2,...,k$, our analysis of Theorem~\ref{Unsupervised-Fashion} on $\mathcal{Q}$ holds.
\end{proof}

\subsubsection*{B2 Proof of Theorem~\ref{thm:kernel mean}}
\begin{proof}
Let $\{\sqrt{\lambda_i}\phi_i(x)\}$ be an orthogonal base of $\mathcal{H}$, the kernel function can be written as: 
\begin{equation}
    K(x,y)=\sum_{i}^{\infty}\lambda_i\phi_i(x)\phi_i(y).
\end{equation}
For any vector $k_x=K(x,\cdot)\in\mathcal{H}$, it can be expressed as a sum of orthogonal bases, \textit{i.e.},  $K(x,\cdot)=\sum_{i}^{\infty}\sqrt{\lambda_i}\phi_i(x)\sqrt{\lambda_i}\phi_i(\cdot)$.
May wish to assume the kernel mean $\mu_{\mathcal{H}}=\sum_{i}^{\infty}\alpha_i\sqrt{\lambda_i}\phi_i(\cdot)$, for any $k_x\in\mathcal{H}$, there exists
\begin{equation}\label{equ:new dh}
    d_{\mathcal{H}}(k_x,\mu_{\mathcal{H}})=\sqrt{\sum_{i}^{\infty}(\sqrt{\lambda_i}\phi_i(x)-\alpha_i)^2}.
\end{equation}
Substituting Eq.~(\ref{equ:new dh}) into Eq.~(\ref{equ:Kernel Fréchet mean}), the kernel mean is equivalent to minimizing the following problem:
\begin{equation}\label{equ:new kernel mean}
           \min_{\mu \in \mathcal{H}} \int \sum_{i}^{\infty}(\sqrt{\lambda_i}\phi_i(x)-\alpha_i)^2 \dif \mathbb{P}(x). 
\end{equation}
The solution of Eq.~(\ref{equ:new kernel mean}) is $\alpha_i=\int\sqrt{\lambda_i} \phi_i(x)\dif \mathbb{P}(x)$, namely, $\mu_{\mathcal{H}}=\sum_{i}^{\infty}\int\sqrt{\lambda_i} \phi_i(x)\dif \mathbb{P}(x)\sqrt{\lambda_i}\phi_i(\cdot)$. To simplify $\mu_{\mathcal{H}}$, we next analyze the interchange of limit and integral. Let $f_t(x)=\sum_i^{t}\sqrt{\lambda_i}\phi_i(x)$, and then $\{f_t(x)\}$ is a Cauchy sequence in $\mathcal{H}$. Since $\mathcal{H}$ denotes the complete metric space, there exists $\lim_{t \to \infty}  f_t(x) =f(x):=\sum_i^{\infty}\sqrt{\lambda_i}\phi_i(x)\in \mathcal{H}$, \textit{i.e.}, 
\begin{equation}
    \lim_{t \to \infty}\| f_t-f \|_{\mathcal{H}}=0.
\end{equation}
The above equation shows that $f_t(x)$ converges to $f(x)$ in norm $\|\cdot\|_{\mathcal{H}}$. For RKHS, the evaluation functional $\delta_x:\delta_x f \mapsto f(x)$ is a continuous linear functional, which means that for all $x\in\mathcal{X}$:
\begin{equation}
\begin{split}
    \lim_{t \to \infty} |f_t(x)-f(x)|  &\leq \lim_{t \to \infty} |\delta_x f_t - \delta_x f| \\
    &\leq \lim_{t \to \infty} \| \delta_x \| \| f_t -f \|_{\mathcal{H}} =0.\\
\end{split}
\end{equation}
The above inequality shows that convergence in norm implies pointwise convergence in RKHS. We now consider the difference between the integral of $f_t(x)$ and the integral of $f(x)$. According to the linearity and monotonicity of the Bochner integral, there exists 
\begin{equation}
    \left|\int f_t(x)\dif\mathbb{P}(x)\!-\!\int f(x)\dif\mathbb{P}(x)\right|\leq\int \left|f_t(x)\!-\!f(x)\right|\dif\mathbb{P}(x).
\end{equation}
By the reverse of Fatou's lemma related to $|f_t(x)-f(x)|$, we have the following inequality:
\begin{equation}
\begin{split}
   & \limsup_{t \to \infty} \int |f_t(x)-f(x)|\dif\mathbb{P}(x)\\ 
    &\leq \int \limsup_{t \to \infty} |f_t(x)-f(x)| \dif\mathbb{P}(x)
      =0,
    \end{split}
\end{equation}
which implies that the limit exists and 
\begin{equation}\label{equ:limit of ft}
   \begin{split}
  &  \lim_{t \to \infty} \left|\int f_t(x)\dif\mathbb{P}(x) -\int f(x)\dif\mathbb{P}(x) \right|\\
    &\leq \lim_{t \to \infty} \int |f_t(x)-f(x)|\dif\mathbb{P}(x) =0.\\
       \end{split}
\end{equation}
The above inequality shows that limit and integral is interchangeable with respect to kernel mean $\mu_{\mathcal{H}}$. Based on the uniqueness of limits for $\mathcal{H}$, the kernel mean is unique with the closed-form solution: 
\begin{equation}
\begin{split}
    \mu_{\mathcal{H}}&=\sum_{i}^{\infty}\int\sqrt{\lambda_i} \phi_i(x)\dif \mathbb{P}(x)\sqrt{\lambda_i}\phi_i(\cdot)\\
    &=\lim_{t \to \infty}\sum_{i}^{t}\int\sqrt{\lambda_i} \phi_i(x)\dif \mathbb{P}(x)\sqrt{\lambda_i}\phi_i(\cdot)\\
    &=\lim_{t \to \infty}\int\sum_{i}^{t}\sqrt{\lambda_i} \phi_i(x)\dif \mathbb{P}(x)\sqrt{\lambda_i}\phi_i(\cdot)\\
    &=\int\lim_{t \to \infty}\sum_{i}^{t}\sqrt{\lambda_i} \phi_i(x)\dif \mathbb{P}(x)\sqrt{\lambda_i}\phi_i(\cdot)\\
    &=\int\sum_{i}^{\infty}\sqrt{\lambda_i} \phi_i(x)\dif \mathbb{P}(x)\sqrt{\lambda_i}\phi_i(\cdot)\\
    &=\int K(x,\cdot) \dif \mathbb{P}(x).
\end{split}
\end{equation}
\end{proof}

\subsection*{Part C: Stein methods and  SVGD algorithm}

\textbf{Stein’s Identity.} 
Given a smooth density  $p(x)$ observed on $\mathcal{X} \subseteq \mathbb{R}^{n}$, let $A_p$ be the Stein operator, $\varphi(x) = {[\varphi_1(x), ..., \varphi_n(x)]}^{\top}$ be the  smooth vector function. For sufficiently regular $\varphi$, Stein’s Identity is defined as:
\begin{equation}\label{eq:SI}
\begin{split}
\mathbb{E}_{x\sim p}[A_p \varphi(x)] = 0,
\end{split}
\end{equation}
where
\begin{equation}
\begin{split}
A_p \varphi(x) = {\varphi(x)}\nabla_x {\rm log} p(x)^{\top} + \nabla_x \varphi(x).
\end{split}
\end{equation}
Here $x \sim p(x)$ denotes the continuous random variable or parameter sampled from $\mathcal{X}$, $\nabla_x \varphi(x) $ denotes the score function of $\varphi(x)$, $Q=A_p \varphi(x)$ denotes the Stein operator $A_p$ acting on function $\varphi$. \\
\textbf{Kernelized Stein Discrepancy.} 
Stein Discrepancy is a discrepancy measure which can maximize violation of Stein’s Identity and can be leveraged to define how different two smooth densities $p$ and $q$ are:
\begin{equation}\label{eq:SD}
{{\rm SD}(q, p)} = {\max \limits_{\varphi \in \mathcal{F}} \Bigg\{\Big[\mathbb{E}_{x\sim q}\Big({\rm trace}(Q)\Big)\Big]^2 \Bigg\}}, 
\end{equation}
where $\mathcal{F}$ denotes a set of smooth functions with bounded Lipschitz norms, it determines the discriminative capability of Stein Discrepancy. However, with \cite{liu2019stein,gorham2015measuring}, it is computationally intractable to resolve the functional optimization problem on $\mathcal{F}$. To bypass this and provide a close-form solution, Liu et al. \cite{liu2016kernelized} introduce Kernelized Stein Discrepancy (KSD) which maximizes $\varphi$ in the unit ball of a reproducing kernel Hilbert space (RKHS) ${\mathcal{H}}$. The definition of KSD is thus presented:
\begin{equation}
\begin{split}\label{eq:KSD}
{{\rm KSD}(q, p)} = &{\max \limits_{\varphi \in {\mathcal{H}}}} \Bigg\{ \Big[\mathbb{E}_{x\sim q}\Big({\rm trace}(Q)\Big)\Big]^2 \Bigg\} \\
 &{\rm s.t.} \quad {\|\varphi\|_{\mathcal{H}} \leqslant 1}.
\end{split}
\end{equation}
With \cite{chwialkowski2016kernel,liu2016kernelized}, given a positive definite kernel $k(x, x^{’}): \mathcal{X} \times \mathcal{X} \to \mathbb{R}$, the optimal solution $\varphi(x)$ of Eq.~(\ref{eq:KSD}) is given as:
\begin{equation}\label{eq:optimal-solution}
\begin{split}
\varphi(x) = {\varphi_{q, p}^*(x) / {\|{\varphi^*_{q, p}}\|}_{\mathcal{H}}},
\end{split}
\end{equation}
where $\varphi_{q, p}^*(\cdot) = \mathbb{E}_{x\sim q}[A_p k(x, \cdot)]$ indicates the optimal direction for gradient descent.
\par With notions of SD and KSD, Liu et al. rethink the goal of variational inference which is defined in Eq.~(\ref{eq:vi-kl}), they consider the distribution set $\Omega$ can be obtained by smooth transforms from a tractable reference distribution $q_0(x)$ where $\Omega$ denotes the set of distributions of random variables which takes the form $r = T(x)$ with density:
\begin{equation}\label{eq:density}
\begin{split}
q_{[T]}(r) = q(\mathcal{R}) \cdot |\det(\nabla_r \mathcal{R})|,
\end{split}
\end{equation}
where $T: \mathcal{X} \to \mathcal{X}$ denotes a smooth transform,  $\mathcal{R}=T^{-1}(r)$ denotes the inverse map of $T(r)$ and $\nabla_r \mathcal{R}$ denotes the Jacobian matrix of $\mathcal{R}$. With the density, there should exist some restrictions for $T$ to ensure the variational optimization in Eq.~(\ref{eq:vi-kl}) feasible. For instance, $T$ must be a one-to-one transform, its corresponding Jacobian matrix should not be computationally intractable. Also, with \cite{rezende2015variational}, it is hard to screen out the optimal parameters for $T$.
\par Therefore, to bypass the above restrictions and minimize the KL divergence in Eq.~(\ref{eq:vi-kl}), an incremental transform ${T(x) = x + \varepsilon \varphi(x)}$ is introduced, where $\varphi(x)$ denotes the smooth function controlling the perturbation direction and $\varepsilon$ denotes the perturbation magnitude. With the transform $T(x)$ and the density defined in Eq.~(\ref{eq:density}), we have ${q_{[T^{-1}]}(x) = q(T(x))\cdot |\det(\nabla_x T(x))|}$ and ${{\rm KL}(q_{[T]} || p) = {\rm KL}(q || p_{[T^{-1}]})}$. We thus have
\begin{equation}
\begin{split}
\nabla_{\varepsilon} {\rm KL}(q || p_{[T^{-1}]}) = \nabla_{\varepsilon} \Big[\int q(x) \log \frac{q(x)}{p_{[T^{-1}]}(x)}\dif x{\Big]}.
\end{split}
\end{equation}
By converting the formula of ${{\rm KL}(q || p_{[T^{-1}]})}$ to the corresponding mathematical expection form, ${\nabla_{\varepsilon} {\rm KL}(q || p_{[T^{-1}]})}$ becomes ${-\mathbb{E}_{x\sim q} {\Big[}{\nabla_{\varepsilon}\log\Big(}p(T(x)) |{\det(\nabla_x T(x))}|{\Big)}{\Big]}}$. Deriving this formula further, we have
\begin{equation}
\begin{split}
\nabla_{\varepsilon} {\rm KL}(q || p_{[T^{-1}]}) = -\mathbb{E}_{x\sim q} \Big[\mathcal{C} + \mathcal{D}{\Big]},
\end{split}
\end{equation}
where 
\begin{equation}
\begin{split}
&\mathcal{C}=(\nabla_{T(x)}\log p(T(x)))^{\top}{\nabla_{\varepsilon} T(x)}, \\
&\mathcal{D}={(\nabla_{x} T(x))^{-1} \nabla_{\varepsilon} \nabla_{x} T(x)}.
\end{split}
\end{equation}
Besides, if ${\varepsilon = 0}$, there exists ${T(x)=x}$, ${\nabla_{\varepsilon}T(x)=\varphi(x)}$, ${\nabla_{x}T(x)=I}$, and ${\nabla_{\varepsilon} \nabla_{x} T(x)=\nabla_{x} \varphi(x)}$. With these knowledge, we present the following theorem.
\begin{theorem}\label{Stein-KL}
Let $T(x) = x + \varepsilon \varphi(x)$, $q_{[T]}(r)$ be the density $r = T(x)$, then
\begin{equation}
\begin{split}
\nabla_{\varepsilon}{\rm KL}(q_{[T]} \| p)|_{\varepsilon=0} =
-\mathbb{E}_{x\sim q}[{\rm trace}(Q)],
\end{split}
\end{equation}
where $\nabla_{\varepsilon}{\rm KL}(q_{[T]} || p)$ denotes the directional derivative of ${\rm KL}(q_{[T]} \|p)$ when $\varepsilon$ tends to be infinitesimal (\textit{i.e.}, $\varepsilon \to 0$).
\end{theorem}
Relating the definition of KSD in Eq.~(\ref{eq:KSD}), Eq.~(\ref{eq:optimal-solution}), and Theorem \ref{Stein-KL}, the optimal perturbation direction for Eq.~(\ref{eq:optimal-solution}) can be identified as  $\varphi^*(\cdot)_{q, p}$. We thus present Lemma \ref{Stein-varphi}.
\begin{lemma}\label{Stein-varphi}
Given the conditions in Theorem \ref{Stein-KL} and consider all the perturbation directions $\varphi$ in the ball $\mathcal{B} = \{\varphi \in \mathcal{H}^n : \|\varphi\|^2_{\mathcal{H}} \leqslant {\rm KSD}(q, p) \}$ 
of $\mathcal{H}^n$ (\textit{i.e.}, RKHS), the direction of steepest descent that maximizes the negative gradient in Eq.~(\ref{eq:optimal-solution}) is $\varphi^*_{q, p}$:
\begin{equation}\label{eq:varphi}
\begin{split}
\varphi^*_{q, p} = \mathbb{E}_{x\sim q}[k(x, \cdot)\nabla_x{\rm log} p(x) + \nabla_x k(x, \cdot)],
\end{split}
\end{equation}
for which we have $\nabla_{\varepsilon}{\rm KL}(q_{[T]} || p)|_{\varepsilon=0}$ = $-{{\rm KSD}(q, p)}$.
\end{lemma}
\end{document}